\documentclass{article}
\usepackage[preprint]{NeurIPS/neurips_2025}  %uncomment for camera ready
\usepackage{comment}

\makeatletter
\newcommand{\anonymize}[2]{%
  \if@submission
    {#1}%
  \else
    {#2}%
  \fi
}
\makeatother

\usepackage{amsmath}
\usepackage[utf8]{inputenc} % allow utf-8 input
\usepackage[T1]{fontenc}    % use 8-bit T1 fonts
\usepackage{hyperref}       % hyperlinks
\usepackage{url}            % simple URL typesetting
\usepackage{booktabs}       % professional-quality tables
\usepackage{amsfonts}       % blackboard math symbols
\usepackage{nicefrac}       % compact symbols for 1/2, etc.
\usepackage{microtype}      % microtypography
\usepackage{xcolor}         % colors
\usepackage{graphicx}
\usepackage{natbib}
\usepackage{todonotes}
\usepackage{marvosym}
\usepackage{float}
\usepackage{caption}
\DeclareCaptionType{Video}[Video]

\DeclareCaptionType{AppendixFig}[Figure][List of Appendix Figures]

\DeclareCaptionLabelFormat{AppendixFigLabel}{#1~#2}
\usepackage[skip=4pt plus1pt,]{parskip}

\newcommand{\x}{\mathbf{x}} % image
\newcommand{\aug}{\mathcal{A}} % augmentation
\newcommand{\z}{\mathbf{z}} % embedding vector
\newcommand{\p}{\mathbf{p}} % projection vector

\title{DynaCLR: Contrastive Learning of Cellular Dynamics with Temporal Regularization}
\author{
\\
Eduardo Hirata-Miyasaki\textsuperscript{1,*},
Soorya Pradeep\textsuperscript{1,*}, 
Ziwen Liu\textsuperscript{1,*}, 
Alishba Imran\textsuperscript{1,2,}\thanks{equal contribution}, \\
Taylla Milena Theodoro\textsuperscript{1}, 
Ivan E. Ivanov\textsuperscript{1}, 
Sudip Khadka\textsuperscript{1}, \\
See-Chi Lee \textsuperscript{1}, 
Michelle Grunberg \textsuperscript{1},
Hunter Woosley\textsuperscript{1},
Madhura Bhave\textsuperscript{1}, \\
Carolina Arias\textsuperscript{1}, 
Shalin B. Mehta\textsuperscript{1,}\thanks{correspondence: \texttt{shalin.mehta@czbiohub.org}}\\
\\
\textsuperscript{1} Chan Zuckerberg Biohub San Francisco, San Francisco, CA 94158, USA\\
\textsuperscript{2} University of California Berkeley, Berkeley, CA 94720, USA\\
}

\begin{document}
% contrastive learning of representations of dynamic cellular responses to perturbations from time-lapse

\maketitle

\begin{abstract}
We report DynaCLR, a self-supervised method for embedding cell and organelle \textbf{Dyna}mics via \textbf{C}ontrastive \textbf{L}earning of \textbf{R}epresentations of time-lapse images. DynaCLR integrates single-cell tracking and time-aware contrastive sampling to learn robust, temporally regularized representations of cell dynamics. DynaCLR embeddings generalize effectively to in-distribution and out-of-distribution datasets, and can be used for several downstream tasks with sparse human annotations. We demonstrate efficient annotations of cell states with a human-in-the-loop using fluorescence and label-free imaging channels. DynaCLR method enables diverse downstream biological analyses: classification of cell division and infection, clustering heterogeneous cell migration patterns, cross-modal distillation of cell states from fluorescence to label-free channel, alignment of asynchronous cellular responses and broken cell tracks, and discovering organelle response due to infection.  DynaCLR is a flexible method for comparative analyses of dynamic cellular responses to pharmacological, microbial, and genetic perturbations. We provide PyTorch-based implementations of the model training and inference pipeline \anonymize{}{(\href{https://github.com/mehta-lab/viscy}{VisCy})}and a GUI \anonymize{}{(\href{https://github.com/czbiohub-sf/napari-iohub}{napari-iohub})} for the visualization and annotation of trajectories of cells in the real space and the embedding space.

% alternative names: cytodyn, DynaCLR, CLRcell

\end{abstract}

\section{Introduction}

% Why: writing for CV audience. Start with data and analysis goals.
Learning biologically interpretable representations of the cell morphology and architecture from 100 TB-scale dynamic imaging datasets is an outstanding need in basic biology and therapeutic discovery. The dynamic responses of organelles and cells to perturbations such as infection, gene expression modulation, or pharmacological treatment can reveal biomarkers of health and disease and establish causal links between the cell morphology and function. Supervised approaches for analyzing dynamic cell morphology are suboptimal because categorical labeling of continuous changes in cell and organelle morphology is hard. Even when relevant morphological cell states can be categorized, human annotation of 3D movies is very expensive and prone to bias. Self-supervised representation learning offers an unbiased approach for robust and diverse analysis of dynamic imaging data. Using biologically and experimentally relevant pretext tasks for encoding the images of single cells and their trajectories, self-supervised methods has the potential to learn robust embeddings of cell and organelle dynamics, generalize across experimental conditions, disambiguate the relationships between complex perturbations and cellular responses, and enable the discovery of rare cell states.  Current self-supervised embedding methods are not designed to encode multi-channel 3D timelapse datasets or enable efficient application of prior knowledge of (dis-)similarity of cell morphology as pretext loss functions.  

% What: computer vision goals.
We report a method to learn temporally regularized embeddings of multi-dimensional time-lapse datasets, named embedding \textbf{Dyna}mics via \textbf{C}ontrastive \textbf{L}earning of \textbf{R}epresentations (DynaCLR). DynaCLR combines single-cell tracking with cell and time-aware contrastive encoding to learn embeddings of cell and organelle dynamics from multi-channel 3D time-lapse microscopy data. Unlike natural images, microscopy images include diverse channels -- fluorescence channels often encode functional reporters, biosensors, or organelle markers, while label-free channels, e.g., phase contrast, encode information about the cell-cycle stage, cell death, or infection. Together, these channels can capture a rich yet complex spectrum of functional states, including cell division, pathogen replication, immune responses, organelle remodeling, and apoptosis. Disentangling cell states from static snapshots of a cell population is hard due to the heterogeneity of states and the asynchronous perturbations and cellular responses. By learning the dynamics of cell states from time-lapse data, DynaCLR enables robust modeling of heterogeneous behavior via categorical classification of cell states or alignment of asynchronous events from dynamic embeddings.

This paper reports the following methodological innovations to enable quantitative analysis of cell and organelle dynamics in response to perturbations:
\begin{enumerate}
    \item DynaCLR embedding models for mapping the 3D multi-channel images of single cells to a temporally regularized embedding space, i.e., the distance in the embedding space changes smoothly when cell and organelle morphology change smoothly. DynaCLR models generalize to out-of-distribution data acquired with diverse imaging systems and cell types, making the learned embeddings useful for multiple downstream analyses. We compare DynaCLR embeddings with two baseline models: ConvNeXt \citep{liu_convnet_2022} trained with ImageNet and OpenPhenom trained with JUMP-CP datasets \citep{kraus_masked_2023}.

    \item Diverse downstream analyses from DynaCLR embeddings, including, classification of the cell states in the embedding space with efficient human annotation, analyzing dynamics of cell states, aligning heterogeneous timing of perturbations, and discovery of changes in cells and organelles due to a specific perturbation. 
    
    \item A scalable PyTorch implementation for training models on GPU clusters \anonymize{}{(\href{https://github.com/mehta-lab/viscy}{VisCy})} and a GUI for annotating cell states in real and embedding spaces \anonymize{}{(\href{https://github.com/czbiohub-sf/napari-iohub}{napari-iohub})}.

\end{enumerate} 

% What: applications
We evaluate the accuracy of the visual representation learned by our method using metrics specific to the downstream task and metrics agnostic to the downstream task.

% We compare our method with two baseline methods:  supervised time-agnostic semantic segmentation of the infection state and self-supervised time-agnostic contrastive learning. We explore the effect of different temporal sampling strategies on the distribution of morphological states in the embedding space and detecting large changes in cell morphology.

% This is a detail that belongs to methods
% Specifically, we use an anchor image, which is a reference image of a cell at a particular time point; a positive pair, which can either be an augmented version of the anchor image or an image of the same cell at a different time point; and a negative pair, which is an image of a different cell. Our approach involves minimizing the distance in the latent space between the anchor image and the positive pair, thereby encouraging similar representations for the same cell over time or under augmentation. Conversely, we aim to maximize the distance between the anchor and the negative pair to ensure that different cells are distinctly represented. This methodology allows the model to generate a set of feature vectors that encapsulate the state of each cell in latent space, providing valuable insights into the temporal progression of cellular changes during infection.

\section{Background and related work}
\label{sec:background}
% General description of self-supervised learning for videos of natural scenes.

Representation learning of images of cells~\citep{he_masked_2021, kraus_masked_2023} is accelerating our ability to learn biological relationships from images. In parallel,   
learning visual representations of objects and scenes from videos~\citep{wang_unsupervised_2015, denton_unsupervised_2017, sermanet_time-contrastive_2018, qian_spatiotemporal_2021, dave_tclr_2021} has been an active area of computer vision.  Among the self-supervised learning approaches, contrastive learning~\citep{hadsell_dimensionality_2006} offers several advantages: it allows the introduction of prior knowledge of the relationships between the data points as a contrastive loss term~\citep{chen_simple_2020, he_momentum_2020}, it can be used with deterministic or generative models~\citep{aneja_contrastive_2021}, and it enables joint embedding of diverse channels and modalities~\citep{radford_learning_2021}. 

In cell biology, self-supervised models of time-lapse microscopy data have enabled diverse analyses, e.g., analysis of immune response~\citep{wu_dynamorph_2022, shannon_cellplato_2024}, profiling of cell lineages~\citep{soelistyo_learning_2022, ulicna_learning_2023}, phenotyping of plant cells~\citep{marin_zapata_self-supervised_2021}, and dense representations of cell dynamics~\citep{gallusser_self-supervised_2023}. In parallel, contrastive self-supervised models of static snapshots have enabled analyses of cell and organelle states, e.g., diversity of mitochondrial shapes~\citep{natekar_self-supervised_2023} in response to perturbations, detection of cell division~\citep{zyss_contrastive_2024}, and learning correlation between gene expression and morphology~\citep{wang_multi-contrastivevae_2024,senbabaoglu_mosby_2024}. Understanding the mechanisms of most dynamic cell state transitions requires time-resolved measurements~\citep{shakarchy_machine_2024, ulicna_learning_2023}. Among self-supervised learning frameworks, contrastive learning enables flexible use of priors for learning robust embeddings, including temporal priors as demonstrated in this work. DynaCLR learns temporally smooth mapping from videos, and yet allows embedding of snapshots or time-lapse data.

% Infection Biology, Organelle interactions. 

The development of DynaCLR has been driven by the need to analyze the remodeling of cells and organelles in response to viral infection and the cell cycle across multiple microscopes. Viruses exploit the host cell's machinery to produce new virions, reprogramming the structure and function of the organelles and the whole cell. For example, flaviviruses, such as Zika and Dengue, replicate on  Endoplasmic Reticulum (ER) derived membrane compartments\citep{verhaegen_endoplasmic_2024}, leading to morphological changes in the ER and other organelles, as well as the whole cell. Over the years, many authors have investigated the global impact of viral infection on cells and organelles using transcriptomic\citep{gutierrez_single-cell_2022} and proteomic\citep{bojkova_proteomics_2020, hein_global_2023, cook_restructured_2022} methods. The -omics modalities have enabled the discovery of the changes in the molecular states of the cells due to perturbations. However, they do not directly report the dynamic remodeling of organelles and cells. Analyzing cell and organelle dynamics in response to perturbations requires 3D multi-channel (e.g., multiple fluorescent channels, with or without phase) imaging. Extracting biological insights from these datasets requires new methods for learning robust models that map 4D tensors in real space to an embedding space that represents diverse cell types and cell states. 

% Repetition of points made above.
% Earlier work on temporally regularized variational autoencoder (VAE) models~\citep{wu_dynamorph_2022} demonstrated that incorporating weak priors about the temporal smoothness of embeddings leads to models that generalize to unseen data.  Contrastive learning is a flexible method for encoding priors of similarity and distance between objects in a dataset across the dimensions of space, time, perturbations, and channels. Contrastive models also tend to be more parameter-efficient for discriminating cell phenotypes due to the absence of the decoder used in generative models, an essential feature for training models that embed 4D tensors. Considering the above trade-offs, we encode complex cell and organelle morphology with 3D multi-channel live cell imaging and decode the cell states using cell tracking and time-aware contrastive sampling.

% Commenting out the work on benchmark datasets.
% Large-scale benchmark datasets of static images of perturbed cells~\citep{chandrasekaran_jump_2023, chen_chammi_2024} are available. However, benchmark datasets of time-lapse images of perturbed cells are smaller in comparison~\citep{edlund_livecelllarge-scale_2021, antonelli_alfi_2023} due to the challenges of live imaging and annotations outlined earlier.

\section{Method}

\subsection{Data and annotations}
\label{sec:exp_data_ann}
We explore the performance and applications of DynaCLR with five distinct time-lapse datasets: (1) a previously published 2D dataset capturing cell cycle dynamics~\citep{antonelli_alfi_2023}, (2) a previously published 3D dataset of perturbed microglia~\citep{wu_dynamorph_2022}, (3) a 5D dataset representing both infection and cell cycle dynamics, and (4) two 5D datasets acquired at different temporal resolutions that encode infection and SEC61, an endoplasmic reticulum (ER) organelle marker. The details of the acquisition, preprocessing, and annotations are summarized in \autoref{sec:datasets}, and the models are summarized in \autoref{tab:summary-of-models} and \autoref{tab:model-input}.

\begin{figure}[h!]
    \centering
    \includegraphics[width=\textwidth]{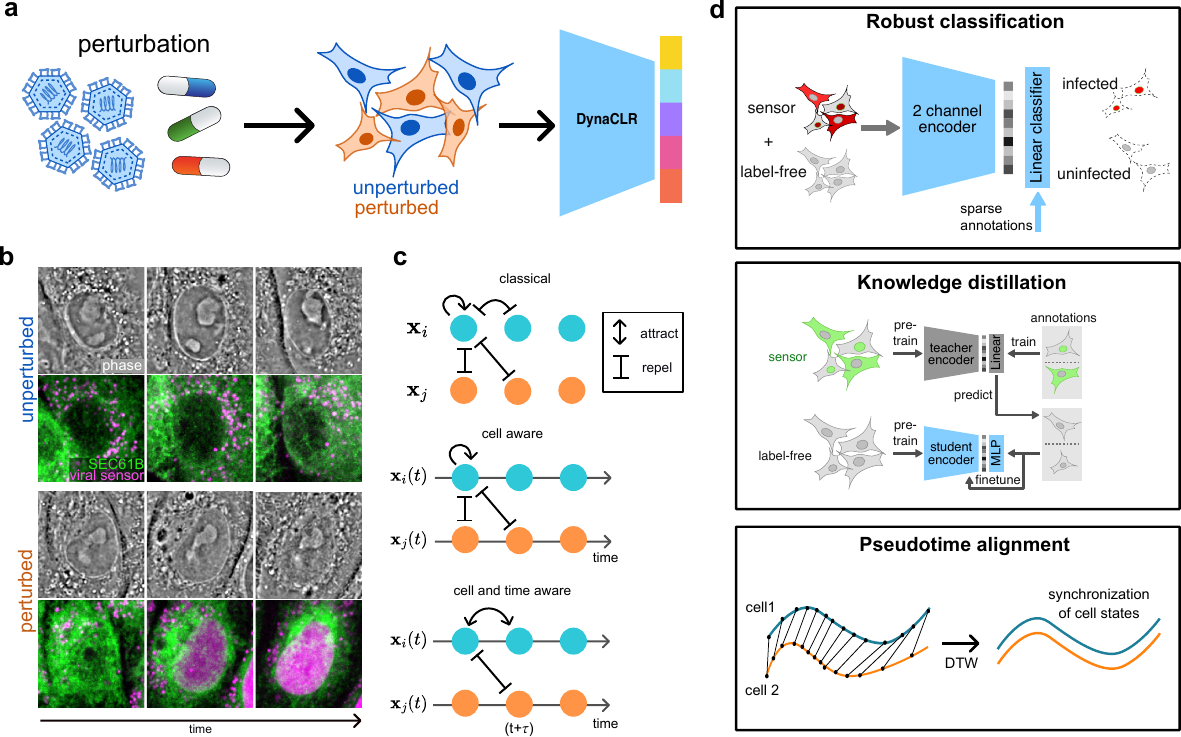}
    \caption{\textbf{Summary of DynaCLR:} (a) Live cells are perturbed, e.g., infected, and time-lapse datasets are acquired at high throughput with correlative quantitative phase and fluorescence microscopy. Cell nuclei are virtually stained and tracked. Single-cell tracks are used to train DynaCLR models. (b) Illustrative patches of unperturbed and perturbed cells imaged with multiple channels: phase (grayscale), viral sensor (magenta), and endoplasmic reticulum marker SEC61 (green). (c) Contrastive loss with three different sampling strategies, classical, cell-aware, and time-aware, is used to map multi-channel volumetric movies to embedding vectors. (d) The learned embeddings enable several downstream tasks: robust classification of multiple cell states with efficient annotations (e.g., infection and cell division), label-free prediction of cell states via cross-modal knowledge distillation, and alignment of trajectories.}
    \label{fig:overview}
\end{figure}

\subsection{Time-aware and cell-aware contrastive sampling}
\label{sec:sampling}
DynaCLR method is illustrated in \autoref{fig:overview}a-c, along with a high-throughput pipeline for live cell imaging, single-cell tracking, and downstream analyses. We embed 3D multi-channel patches of single cells $\x_{i}(t) \in \mathbb{R}^{C\cdot Z\cdot Y\cdot X}$, where C denotes channels and Z, Y, X are spatial dimensions. The cells are subjected to different perturbations, including the intrinsic perturbation of time~(\autoref{fig:overview}b). DynaCLR method works with diverse information channels, including fluorescence channels that report molecular architecture and label-free channels~\footnote{ Note that \emph{label-free} in the context of biological microscopy implies the absence of fluorescent labeling of cells and not necessarily the absence of human annotations of cell states.}  that report physical architecture.
The cells \( \x_i \) are tracked across time \( t_1, t_2, \ldots, t_n \) as they transition through different states, e.g., division, infection, death, and innate immune response. DynaCLR models are trained with a set $\{\x_i(t)\}$ of tracks using different contrastive sampling strategies, where $i$ is the track ID and not a batch index. DynaCLR models ($f$) map the tracks in image space to temporally regularized embeddings \( \z_i(t) = f [\x_i(t)]\). We trained DynaCLR models using either explicit negative sampling and triplet loss~\citep{weinberger_distance_2005} or implicit negative sampling and NT-Xent loss~\citep {chen_simple_2020}. 

We evaluate three sampling strategies (~\autoref{fig:overview}c, \autoref{sec:modelarch}), summarized below, that can flexibly leverage morphological information encoded at diverse spatial and temporal resolutions across multiple channels. 
\begin{itemize}
    \setlength{\itemsep}{0pt}
    \item \textbf{Classical sampling:} This strategy is the same as classical contrastive sampling of natural images and does not use track index or time index to choose positive or negative pairs. 
    %The pretext task embeds two augmented views of an anchor cell at a given time $\x_i(t)$ as positive pairs and all other images of the same cell at different time points or other cells as negative pairs. 
    The anchor $\aug_1[\x_i]$  and positive pair $\aug_2[\x_i]$ are created through augmentations $\aug$ of a cell at given time point, while negative examples are augmented views of random cells $\aug_3[\x_j]$ at random time points.
    
    \item \textbf{Cell aware sampling:} This strategy uses tracking to form the positive pairs from the images of the same cell and negative pairs from the images of distinct cells, when explicit negative sampling is used. When triplet loss is used, the positive pairs $\{(\aug_1[\x_i], \aug_2[\x_i])\}$ are created from augmentations of the anchor image, and the negative pairs $\{(\aug_1[\x_i],\aug_3[\x_j]), i \neq j\}$ are images of other cells at random times.  The triplet loss is computed over batches  $\mathcal{B}=\{(\aug_1[\x_i], \aug_2[\x_i], \aug_3[\x_j]), i \neq j\}$ consisting of anchor, positive, or negative triplets. The NT-Xent loss is computed over batches  $\mathcal{B}=\{(\aug_1[\x_i], \aug_2[\x_i]\}$ of anchor-positive pairs. \emph{When implicit negative sampling with NT-Xent loss is used, cell-aware and classical sampling have the same behavior}, i.e., the pretext task is to minimize the distance between embeddings of augmented views of a given cell at a given time, and simultaneously maximize the embedding distance from all other cells.

    \item \textbf{Time aware sampling:} This strategy uses tracking to sample images of the same cell at $t$ and $t+\tau$ to form positive pairs $\{(\aug_1[\x_i(t)], \aug_2[\x_i(t+\tau)])\}$.  For explicit negative sampling, an image of a different cell $\aug_3[x_j(t+\tau)]$ at time point $t+\tau$ is sampled as a negative example. The triplet loss is computed over the batch of triplets $\mathcal{B}=\{(\aug_1[\x_i(t)], \aug_2[\x_i(t+\tau)], \aug_3[\x_j (t+\tau]), i \neq j\}$. The NT-Xent loss is computed over the batch of pairs  $\mathcal{B}=\{(\aug_1[\x_i(t)], \aug_2[\x_i(t+\tau)])\}$. The pretext task is to minimize the distance between embeddings of a given cell across the time interval $\tau$ and, simultaneously, maximize the distance between the embeddings of other cells over the same time interval. The time offset $\tau$ is a hyperparameter empirically chosen based on the time scales of the dynamic process and time resolution of imaging. For the experiments in this paper, each pair is typically constructed from adjacent frames. We also explore the effect of time offset ($\tau$) on the structure of embeddings in datasets with high temporal sampling.  Time-aware implicit negative sampling is analogous to time-arrow prior~\citep{gallusser_self-supervised_2023}, but in the embedding space rather than pixel space. At inference time, explicit time-arrow prior~\citep{gallusser_self-supervised_2023} requires multiple frames from a video, whereas implicit prior used by DynaCLR allows flexible embedding of static snapshots and videos.
\end{itemize}

The model architecture, training, and data augmentations are described in the \autoref{sec:modelarch} and \autoref{tab:augmentations}. 

\subsection{Analysis of dynamics in embedding space and metrics}
DynaCLR embeddings enable diverse downstream analyses. In this work, we focus on robust cell state classification with efficient human annotation, cross-modal knowledge distillation of cell states from biomarkers to label-free imaging channels, and temporal alignment of asynchronous cell state dynamics. Among these, the alignment task benefits the most from temporally regularized embeddings (\autoref{sec:alignment}). A task-specific metric of classification accuracy, based on expert annotations, is used to evaluate models. In addition, task-agnostic metrics of smoothness and dynamic range of embedding space, adapted from ~\citep{wu_dynamorph_2022} and described in ~\autoref{sec:metrics}, are used to assess that trained models are temporally regularized, but do not overfit on tracks. Throughout this work, out-of-distribution biological experiments are used to evaluate the generalization of the models.

 We developed a \anonymize{napari}{napari-iohub} plugin (\autoref{vid:napari}) to link the dynamics of cells in the embedding space (visualized via UMAP, PHATE, or PCA projections) with dynamics in the real space in multiple channels to enable efficient human-in-the-loop annotations. We train lightweight, robust classification heads to detect cell states with efficient annotations (e.g., infection and cell division). We employ a knowledge distillation strategy to train classifiers of cell states from label-free data that are challenging to annotate. 

Lastly, we align the heterogeneous temporal kinetics of perturbations of single cells to analyze the responses. This technique allows the comparison of heterogeneous cellular responses across different time points and experimental conditions. We used dynamic time warping (DTW) (\citep{sakoe_dynamic_1978, ulicna_learning_2023}) to align asynchronous trajectories of single-cell embeddings, revealing conserved sequences of morphological changes during infection progression. Our implementation is described in \autoref{sec:dtw}.

\section{Experiments}
\label{experiments}

\subsection{Temporal regularization via time-aware contrastive sampling}
\label{sec:temporal_regularization}

We evaluated the effect of contrastive sampling strategies on the smoothness and structure of the learned embeddings using the ALFI dataset~\citep{antonelli_alfi_2023}, which provides annotated mitotic events with 7-minute temporal resolution. We call this set of models, DynaCLR-ALFI~\autoref{tab:summary-of-models}. We compared classical, cell-aware, and time-aware sampling strategies, trained with both triplet and NT-Xent losses (\autoref{tab:app_sampling_strategies_alfi}). 

Time-aware sampling improved temporal continuity and dynamic range of embeddings (\autoref{tab:app_sampling_strategies_alfi}). These improvements are evident from the PHATE projections of test cells (\autoref{fig:contrastive-sampling}b–c, \autoref{fig:alfi-PHATE}), and from the distribution of distances between pairs of frames in embedding space~\autoref{fig:alfi-cosine}. DynaCLR embeddings have better smoothness and dynamic range (\autoref{tab:app_sampling_strategies_alfi}) relative to embeddings from models not trained on the ALFI dataset(\autoref{fig:imagenet-pretrained-ALFI}), especially when time-aware sampling is used.

\subsubsection{Embedding cell division dynamics}
\begin{figure}[h!]
    \centering
    \includegraphics[width=\textwidth]{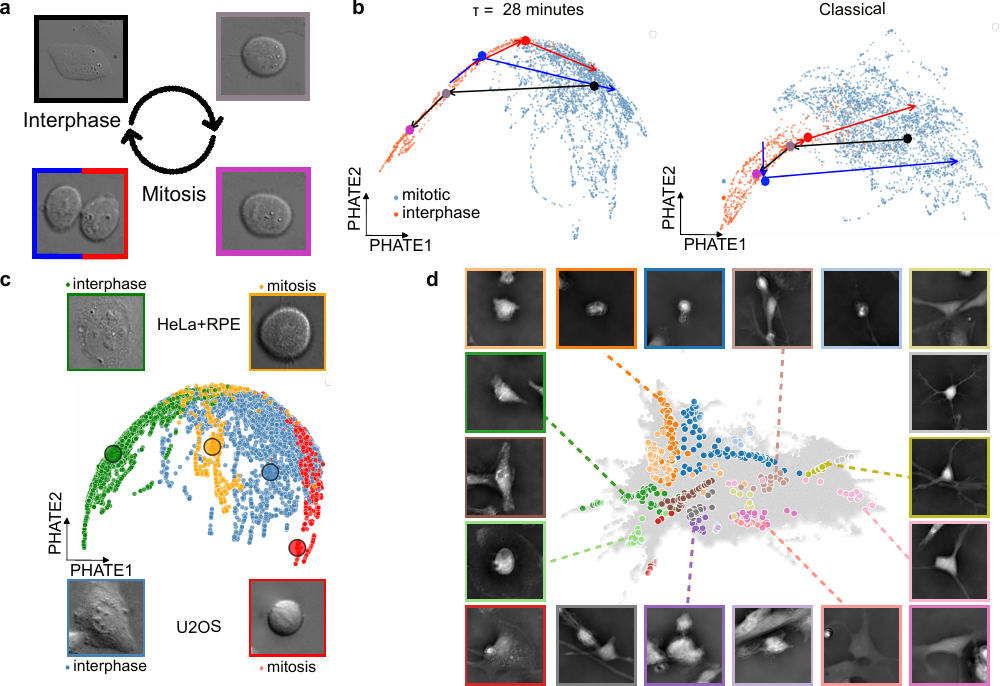}
    \caption{\textbf{Temporally regularized embeddings of cell division and cell migration:} (a) An example track of a parent cell in interphase then undergoing mitosis to form two daughter cells. (b) Trajectories of the dividing cell are displayed using PHATE maps as the cell transitions from the interphase cluster (parent, black track) to the mitosis cluster, followed by the daughter cells (red and blue tracks) re-entering interphase clusters for a time-aware model sampled at 28 minutes and a classical model. (c) Train (U2OS cells) and test (HeLa + RPE1 cells) sets jointly embedded in a PHATE map to show the clustering of cells based on cell cycle state and cell size. (d) PHATE map of microglia morphotypes from a brain tumor environment (glioblastoma). Randomly chosen tracks are highlighted with distinct color-coded points. The images of cells sampled from these tracks (border color indicates the track) illustrate consistent embedding of similar cell morphologies. 
 }
\label{fig:contrastive-sampling}
\end{figure}
First, we focused on cell division to evaluate how well DynaCLR embeddings capture cell state transitions.  To evaluate how these embeddings generalize, we projected U2OS, HeLa, and RPE1 cells from the ALFI dataset into a shared space using a linear classifier trained on cell division (\autoref{fig:contrastive-sampling}c). Notably, the embeddings separated cells by type and state according to the similarity of their shapes and difference in size(\autoref{fig:contrastive-sampling}c). 

% DynaCLR cell division model generalizes to unseen cell types because of a more difficult pretext task -- predicting morphology across time -- leading to temporally consistent features that capture the progression of cell-state dynamics. 

\subsubsection{Embedding immune cell migration dynamics}
Next, we evaluated the ability of the DynaCLR method to generalize across complex and heterogeneous cell morphology exhibited by migratory immune cells. The published DynaMorph embeddings~\citep{wu_dynamorph_2022}, learned using a temporally regularized variational autoencoder, served as a baseline. We trained a DynaCLR-microglia model on cells exposed to pharmacological and infection stimuli (IL-17, IF-$\beta$, Rubella). We used the model to embed an out-of-distribution test set of microglia exposed to a complex disease stimulus (extract from brain tumor glioblastoma). PHATE visualizations in~(\autoref{fig:contrastive-sampling}e) show that the heterogeneity of cell morphology and dynamics is visibly better discriminated than the embeddings learned with a temporally regularized VQ-VAE model~\citep{wu_dynamorph_2022}. This experiment illustrates that temporally regularized contrastive loss improves the separation among heterogeneous cell morphologies and dynamics.

\subsection{Robust classification with DynaCLR embeddings with a human in the loop}
\label{sec:generalization}
% \subsection{DynaCLR models generalize across experiments and microscopes}
% 

\begin{figure}[h!]
    \centering
    \includegraphics[width=1.0\linewidth]{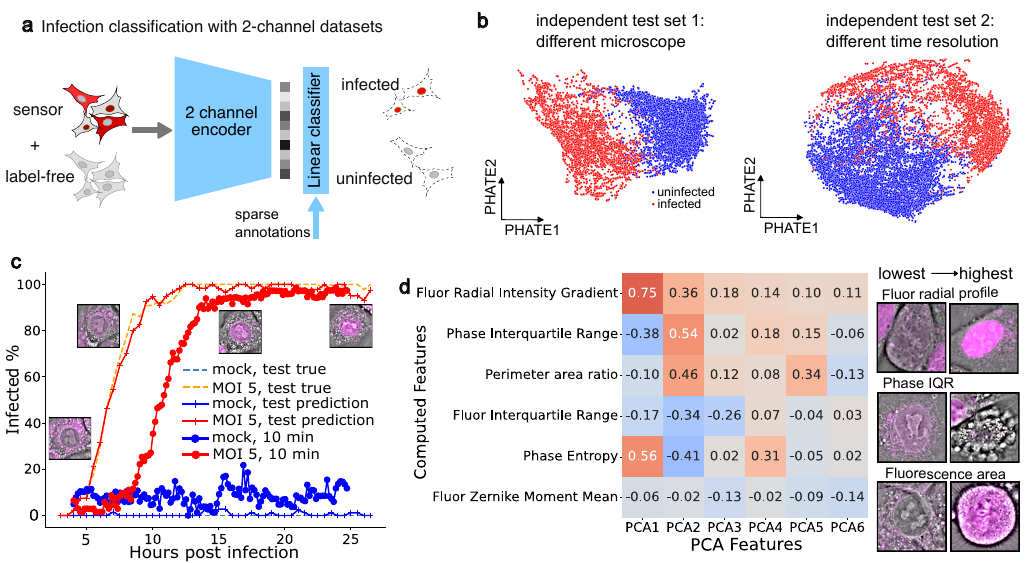}
    \caption{\textbf{DynaCLR embeddings generalize across cell types and microscopes:} a) Viral sensor (fluorescence) and label-free channels are used to train DynaCLR-DENV-VS+Ph model. The resulting embeddings are passed to a linear classifier trained with sparse annotations to predict the infection state of individual cells. b) PHATE visualizations demonstrate the generalization of DynaCLR embeddings to: (1) independent test dataset acquired from a different microscope and (2) data with higher temporal resolution (10 min). Clear separation between infected and uninfected states illustrates robust generalization. c) Infection progression over time is quantified across datasets shown in (b), revealing a steady increase in the percentage of infected cells. d) Correlation analysis between selected morphological features and principal components (PCs) of the embeddings. Representative image patches highlight morphological differences associated with infection states.
} 
    \label{fig:generalization-infection}
\end{figure}

We used phase and viral sensor channels from Dengue virus infection (multiplicity of infection, MOI = 5, i.e., ~5 viral particles per cell) and mock (no virus, MOI = 0) to train DynaCLR-DENV-VS+Ph model for infection phenotyping (\autoref{fig:generalization-infection}a). The embeddings were classified with a linear classifier trained only on a few annotations for infection state and cell division. \autoref{fig:generalization-infection}b shows the PHATE visualization of the embeddings of the test datasets from a different microscope and with a different time resolution, with an overlay of the predicted class. The computed percentage of infected cells from half of the test data closely matched the infection percentages derived from human-revised infection dynamics in mock and MOI 5 conditions, with the number of infected cells rising exponentially and plateauing at 12 hours post-infection (HPI)(\autoref{sec:bio_n_imaging_methods}). A similar trend was observed in the independent test data, where infections plateaued at 15 HPI (\autoref{fig:generalization-infection}c). Thus, the infection classification model trained with DynaCLR method demonstrated robust generalization across microscopes and multiple experiments.

We used the same collection of embeddings to assess robust detection of cell division in uninfected and infected cells(\autoref{fig:cell-division-a549}).  Time-regularized DynaCLR embeddings show smoother transitions as cells transition from interphase to mitosis, as seen from the tracks in the UMAP space (\autoref{fig:cell-division-a549}c and e,\autoref{fig:imagenet-pretrained-ALFI}). 
In contrast, models without temporal regularization produced noisier, random-walk trajectories (\autoref{fig:cell-division-a549}d and f).

We compared the above models to ImageNet- and OpenPhenom-pretrained models, using them only with a single fluorescence channel (viral sensor). While these pretrained models achieve similar F1 scores (\autoref{fig:imagenet-pretrained-infection}) for infection state classification, they lack temporal consistency in the embedding space compared to DynaCLR.

\subsubsection{Explanations of the robust classification of infection and cell division states}
We explored explanations of the phenotypes learned by the DynaCLR model trained on infected cells using two approaches: (a) rank correlation between principal components of the learned embeddings and engineered features, and (b) feature attribution.

Rank correlations between PCs and computed cell features revealed which features were most strongly associated with infection states (\autoref{fig:generalization-infection}d). For example, PC1 captured radial redistribution of fluorescent components and cellular optical heterogeneity (fluorescence radial intensity, phase entropy), while PC2 reflected morphological changes with compartmentalized fluorescence distribution (phase IQR, perimeter-to-area ratio, negative fluorescence IQR correlation), respectively (\autoref{fig:generalization-infection}d). In addition, visual inspection of cell patches along PC axes explain the variations in cell and organelle morphology in the embedding space (\autoref{fig:generalization-infection}d). Detailed correlations between principal components and image features are provided in (~\autoref{fig:PC_vs_CF_sensor}). These correlations demonstrate that the model is sensitive to changes in image features relevant to biology, i.e., changes in the localization of viral sensor and roughness of cell density.

To identify the image patterns that drive cell state classification, we use the occlusion-based feature attribution from Captum \citep{kokhlikyan_captum_nodate}, which highlights the most influential input pixels to the predictions \citep{fleet_visualizing_2014}. Classification heads for infection and division states are attached to the same encoder trained with phase and viral sensor channels and time-and-cell-aware sampling.
The computed attribution maps (\autoref{fig:explain-embeddings}) reveal that the encoder, through self-supervised training, learns meaningful features that describe cell state dynamics, such as viral sensor translocation for infection and chromosome condensation for division. 

\subsection{Label-free classification of infection via cross-modal knowledge distillation}
Robust classification of the functional states of cells, such as infection state, from label-free images can accelerate dynamic phenotyping and enable imaging of organelles in fluorescence channels.
However, human annotation of cell states in label-free time-lapse images is not scalable. It is often possible to acquire paired label-free images with a molecular reporter of cell state. In this section, we report a cross-modal knowledge distillation strategy using DynaCLR embeddings of a fluorescent reporter and phase images for label-free cell state classification. The experiments we report here enable label-free classification of Zika virus (ZIKV) infection. These models enable live cell image-based screens of cellular remodeling due to ZIKV infection and discovery of actionable therapeutics that may inhibit the infection.

\begin{figure}[h!]
    \centering
    \includegraphics{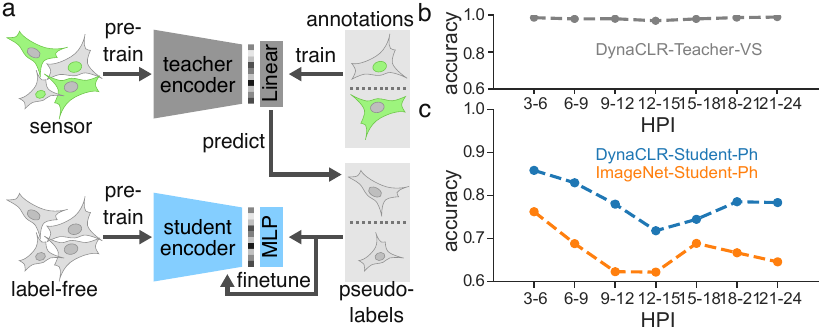}
    \caption{\textbf{Annotation-efficient classification via knowledge distillation}
        (a) Using the DynaCLR method, a teacher model (DynaCLR-Teacher-VS)
        is pretrained on the viral sensor images.
        A linear classifier is then trained with a small human-annotated dataset. The teacher model is used to pseudo-label cell state in a larger dataset with paired fluorescence and label-free channels. 
        A student model (DynaCLR-Student-Ph) is pretrained with label-free images
        and finetuned with pseudo-labels generated by the teacher model using the fluorescence channel.
        (b) Linear classification accuracy of DynaCLR-Teacher-VS on the pseudo-labeling dataset evaluated using additional annotation across the course of infection.
        (c) The classification accuracy of DynaCLR-Student-Ph is consistently higher than the baseline
         model pretrained with ImageNet dataset (ImageNet-Student-Ph) across the course of infection.
    }
    \label{fig:knowledge-distillation}
\end{figure}

An accurate linear classification head can be trained with a small (10,510 instances) annotated dataset (\autoref{sec:generalization}) using DynaCLR pretraining with the fluorescence viral sensor images.  
The combined encoder and classifier serves as the `teacher model' (DynaCLR-Teacher-VS) 
that generates pseudo-labels of the infection state for much larger datasets
with paired label-free and fluorescence images (\autoref{fig:knowledge-distillation}a).
Using a teacher model, we obtained 133,214 annotations of infection state for training a `student model`, DynaCLR-Student-Ph,
amplifying the human annotation by an order of magnitude with minimal loss of accuracy (\autoref{fig:knowledge-distillation}b).
DynaCLR-Student-Ph was first pretrained just with phase images, and finetuned end-to-end with pseudo-labels to classify infection state from phase images (\autoref{fig:knowledge-distillation}a).
The DynaCLR-pretraining consistently outperforms the ImageNet-pretrained feature extraction model (convnextv2\_tiny) (\autoref{fig:knowledge-distillation}c) throughout infection, demonstrating that DynaCLR pertaining improves the model's ability to decode cell state dynamics.
This illustrates that combining fluorescence and label-free imaging, DynaCLR embedding, and cross-modality knowledge distillation
further reduces the burden of human annotation of complex cellular dynamics.

\subsection{Organelle remodeling during infection}
\label{sec:organelle-remod-infection}
Viral infection causes restructuring of organelles, such as the condensation of the ER where replication sites are established \citep{cortese_integrative_2020, scherer_fluorescent_2021}. The range of organelle responses to specific perturbations can be challenging to define a priori. By tracking cells in the learned representation space, we can correlate the observed organelle remodeling with other cell states, such as infection and cell cycle. 

\begin{figure}[h!]
    \centering
    \includegraphics[width=\textwidth]{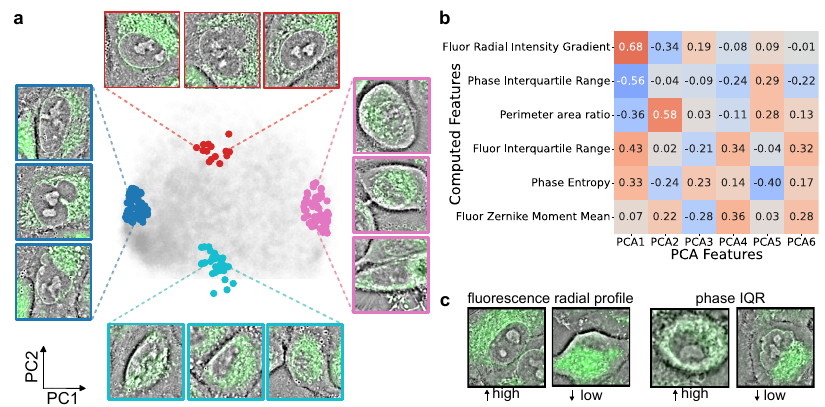}
    \caption{
        \textbf{Discovering cell and organelle responses}:
        (a) Embedding annotations of DynaCLR-DENV-ER+Ph in PC1 vs PC2 space showing the corresponding phenotypes overlaying quantitative phase and SEC61 fluorescent marker. (b) Ranked feature correlation of computed cell features and top principal components. The most correlated features are exemplified by (c) using fluorescence radial distribution and quantitative phase inter-quartile range (IQR). 
    }
    \label{fig:organelle-dynamics}
\end{figure}

We trained a time-aware DynaCLR-DENV-ER+Ph to explore these relationships using a 30-minute temporal resolution dataset using the ER fluorescence marker and phase imaging. Structural changes of the cell and the ER are shown in \autoref{fig:organelle-dynamics}a,c, which show progressive ER condensation through infection. We computed ranked correlations between the engineered cell features and the principal components of the learned embeddings (\autoref{fig:organelle-dynamics}a-b, \autoref{fig:PC_vs_CF_organelle}) to interpret the representation space and identify the features that strongly contribute to the observed structure of embeddings. The model reliably captured ER remodeling due to infection within and out of distribution test sets (\autoref{fig:organelle-remodelling-phate}). 

We sought to analyze ER remodeling \emph{due to} cell division and infection. The DynaCLR-DENV-VS+Ph and DynaCLR-DENV-ER+Ph models enabled embedding of 3-channel imaging data (phase + viral sensor + SEC61) for discovery of changes in ER due to infection~\autoref{fig:organelle-dynamics-umap} and cell division. However, asynchronous cell division and infection state transitions posed a challenge for quantitative analysis.

\subsection{Pseudo-time alignment of cell state dynamics using embeddings}
\label{sec:alignment}
Asynchronous cell state dynamics are often interpreted as heterogeneity of functional states. The temporal variability and individual cell clocks complicate population-level analysis of cellular dynamics. To address this, we use dynamic time warping (DTW) to align cell trajectories in embedding space, revealing conserved state transition patterns despite temporal variations \citep{sakoe_dynamic_1978}. Compared to raw or hand-engineered features, temporally regularized and biologically meaningful embeddings learned by DynaCLR enable alignment of cell trajectories robust to noise and variations in experimental noise. 

DTW has been used to align cell cycle \citep{ulicna_learning_2023, leger_sequence_2025} with diverse embedding models. DynaCLR-ALFI model enabled alignment of mitosis events with the ALFI dataset~\autoref{vid:dtw-alfi}~\citep{antonelli_alfi_2023}. 

Using DTW analysis of embeddings computed with DynaCLR-DENV-VS+Ph model, we implemented pseudo-time alignment for the infection state.  We selected representative reference trajectories showing complete infection progression and aligned all query trajectories in the embedding space with DTW to arrive at a pseudo-time infection axis. 

\begin{figure}[h!]
    \centering
    \includegraphics[width=\textwidth]{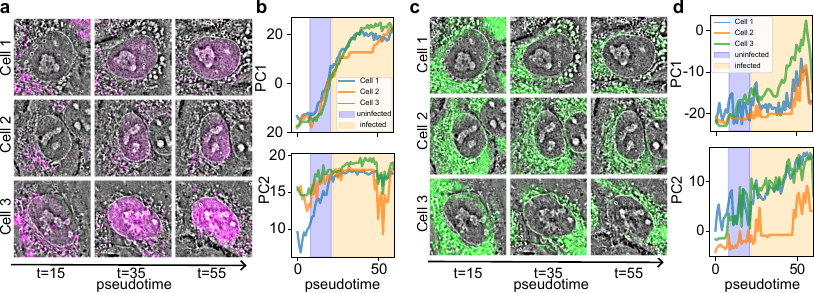}
    \caption{
        \textbf{Temporal alignment of trajectories with learned embeddings}
        (a) Aligned response of three cells during infection, indicated by nuclear translocation of the viral sensor (magenta) in the quantitative phase overlay.(b) Corresponding principal component (PC) trajectories over time showing the uninfected (purple) and infected (yellow) human annotations. (c) Embeddings from the quantitative phase and SEC61, an ER marker (green), are synchronized according to infection state using the temporally regularized embeddings. (d) Corresponding principal component (PC) trajectories of the respective cell embeddings over time.
    }
    \label{fig:organelle_alignment_response}
\end{figure}

Principal component analysis (PCA) of the aligned embeddings reveals consistent trajectories over infection pseudo-time (\autoref{fig:organelle_alignment_response}), highlighting conserved morphological features of progression. This alignment approach enables mapping each cell trajectory to a common pseudo-time axis where similar states align regardless of their real-time occurrence. Despite apparent heterogeneity in real-time measurements, Dengue virus-infected cells follow remarkably similar progression patterns through embedding space when properly aligned (\autoref{fig:organelle_alignment_response}b). 

The temporally regularized DynaCLR embeddings are smoother relative to OpenPhenom embeddings and enable more robust DTW alignment (~\autoref{fig:alignment-model-comparison}). DynaCLR embeddings are also robust to tracking errors in high-density cell culture (~\autoref{fig:bridge-tracks}). Such robustness to the parameters of imaging or downstream analysis is a key advantage of temporal regularization. 

% We could determine whether a drug treatment altered the trajectories of infection or simply accelerated or delayed certain transitions.

\section{Conclusion and future work}
\label{sec:conclusion}
% We will combine the discussion and conclusion in one section.
% The first paragraph should be a zinger that summarizes all main contributions.
These results demonstrate that DynaCLR method leads to a robust temporally regularized embedding space that represents morphological transitions in cells and organelles, and enables diverse downstream analysis tasks: cell state classification, analysis of morphological heterogeneity, cross-channel distillation of cell states, robust alignment of asynchronous cellular trajectories, and analysis of organelle responses to perturbations.

% For this paper:
% NT-Xent
% Finer temporal sampling 

% For future paper:
% Embedding Organelles
% Generalization to public datasets.
The above datasets, computational experiments, and analyses open the following avenues for learning cell state dynamics, among others: 
Learning a foundational model representing diverse cell states in many cell types is an exciting area of research. 
Our current models pair label-free and fluorescence channels to encode cell and organelle states. Training channel-adaptive models that provide biologically interpretable embeddings of datasets with heterogeneous channels is an exciting future direction. The pseudo-time alignment can enable three critical analyses: characterizing the conserved sequence of morphological changes due to perturbations, identifying fate-determining branch points, and measuring the variability in state transition timing across cells.
Predicting the future state of cells (embedding) from the trajectories of embedding can illuminate causal links between the cell states.

\anonymize{}
{
% text for de-anonimized here
\section*{Data and code availability}

The model architecture, training, and prediction code for the DynaCLR method is available at \url{https://github.com/mehta-lab/viscy}. The napari plugin for visualization of data, tracking results, embedding predictions, and performing human annotation is available at \url{https://github.com/czbiohub-sf/napari-iohub}. VisCy is built on PyTorch Lightning, MONAI libraries, and OME-Zarr data format. We used \href{https://github.com/czbiohub-sf/iohub} to convert image data into OME-Zarr format and to load data for training and inference. We used the development version of \href{https://github.com/royerlab/ultrack}{https://github.com/royerlab/ultrack} for single-cell tracking. We used reconstruction algorithms of \url{https://github.com/mehta-lab/waveorder} to compute 3D phase from 3D brightfield volumes.

\begin{ack}
We thank Talon Chandler, CZ Biohub SF, and Sandra Schmid, CZ Biohub SF, for critical feedback on the manuscript. 
The Chan Zuckerberg Initiative funded this research through the Chan Zuckerberg Biohub, San Francisco. All authors are supported by the intramural program of the Chan Zuckerberg Biohub, San Francisco. 
\end{ack}
}

\newpage

\bibliography{RepresentationLearning}
\bibliographystyle{NeurIPS/neurips_2025}

\newpage
\appendix

\begin{center}
  \LARGE\bfseries Appendix
\end{center}
% \section{Appendix}

\section{Datasets}
\label{sec:datasets}
\subsection{ALFI: 2D label-free movies with cell cycle annotations}  We used label-free Differential Interference Contrast (DIC) microscopy movies of three cell types (HeLa, RPE1, and U2OS) from the ALFI dataset by~\citet{antonelli_alfi_2023}. In this dataset, bounding boxes of a subset of cells were tracked and annotated by human experts, with each time point labeled according to the corresponding cell cycle stage (mitosis or interphase). We evaluate the ability of DynaCLR models to discriminate between mitosis and interphase cell cycle stages, which generalizes to unseen cell types. The models were trained with perturbed and unperturbed U2OS cells and tested on unperturbed HeLa and RPE1 cells. All movies were acquired in 2D with a time resolution of 7 min.

\subsection{Microglia: 3D label-free movies of pharmacologically perturbed human microglia} 
This dataset consists of label-free movies of human microglia cells subjected to pharmacological modulators of immune activity  (IL-17, IF-$\beta$, extract of brain tumor glioblastoma) acquired with quantitative phase imaging (QPI) that was used to develop Dynamorph models~\citep{wu_dynamorph_2022}. Dynamorph method used temporally-regularized VQ-VAE and provided a useful baseline to evaluate the generalization, smoothness, and dynamic range of DynaCLR embeddings. For training, we selected movies of cells treated with IL-17, IF-$\beta$, and untreated control conditions, while the glioblastoma supernatant-treated condition was held out for testing. 

Each condition contains nine non-overlapping fields of view (FOV), each containing approximately 250 cells. The raw data were acquired with 1~$\mu$m z-steps every 9 minutes using a Leica DMI-8 microscope with a 20$\times$ objective using QLIPP method~\citep{guo_revealing_2020}. Although the dataset contained bright-field, quantitative phase, and retardance channels, we only used the quantitative phase channel to perform virtual staining of nuclei with CytoLand~\citep{liu_robust_2024} and joint segmentation and tracking with Ultrack~\citep{bragantini_ultrack_2024}, and subsequent model evaluation.

\subsection{Infected cells: 3D label-free and fluorescence movies of Dengue-infected A549 cells}  
\label{sec:bio_n_imaging_methods}
We used 5D time-lapse datasets of A549 cells infected with live Dengue virus to evaluate DynaCLR’s ability to disambiguate dynamic morphological states. The data were acquired using spinning disk confocal and light-sheet microscopes at two temporal resolutions (10 min and 30 min), under both mock (no virus, MOI 0) and infected (MOI 5) conditions. Each movie included a quantitative phase channel and a fluorescence channel encoding infection via a genetically engineered mCherry-NLS sensor~\citep{pahmeier_versatile_2021}. A549 cells with an ER marker, SEC61, infected with Dengue virus, were used to develop the methods for analyzing the organelle remodeling due to the infection.

Cells were segmented using virtual staining~\citep{liu_robust_2024} and tracked with Ultrack~\citep{bragantini_ultrack_2024}.

\subsubsection{Image acquisition and processing}
\label{sec:acq-preproc}
We acquired 5D image datasets (time series of 3D volumes with phase and fluorescence channels) of A549 cells infected with Dengue virus at an MOI of 0 and 5, using:
\begin{itemize}
  \item A spinning disk confocal microscope with 30 min temporal resolution and 0.25 um z-resolution and
  \item A light-sheet microscope with 10-minute and 30-minute temporal resolutions and 0.7 um z-resolution.
\end{itemize}
Mock wells served as controls. Imaging was performed for up to 24 hours in multi-well plates. Image acquisition was automated using \texttt{Micro-Manager}~\citep{edelstein_computer_2010}, and the resulting OME-TIFF files were converted to OME-Zarr using \texttt{iohub} for scalable I/O and downstream processing.

Phase images were reconstructed from brightfield z-stacks using Köhler illumination~\citep{guo_revealing_2019}, and normalized per field-of-view to zero mean and unit variance. Fluorescence images were normalized per field-of-view centered around the median and scaled to range between the 50th and 99th percentile, effectively centering the background at zero while preserving signal dynamic range.

Virtual staining of cell nuclei was performed using a deep learning model~\citep{liu_robust_2024}, followed by segmentation and tracking via Ultrack~\citep{bragantini_ultrack_2024}. 3D image patches of single cells were cropped based on track centroids. We applied extensive augmentations at training time (Table 3) to simulate imaging variability and improve generalization.

Together, these steps encoded the intrinsic perturbations of the cell cycle, the extrinsic perturbations of the infection cycle, and the response of organelles in a dataset of movies suitable for DynaCLR model training.

\subsubsection{Annotations of infection and cell division}
\label{sec:annotation_infection_division}
We evaluated the trained model with a test set curated by an expert who annotated cell division and infection states. We validated the annotations and predictions by overlaying them on the projected embeddings. We also tested the model on independent test data to assess its generalization to new data.

Infection state annotation was based on manually revised annotations from a 2D-Unet model \citep{liu_viscy_2023}, adapted for semantic segmentation and three-class classification using weighted cross-entropy loss. The model classified patches of pixels into three categories: background (0), uninfected nuclei (1), and infected nuclei (2). The annotations were proofread and edited using a custom napari~\citep{chiu_napari_2022} plugin. The proofreading of the semantic segmentation model's predictions was necessary due to the inability to accurately capture late infection stages and cell death, as these states often resulted in a loss of fluorescence signal and altered cell morphology.

Cell division is captured from cell tracking by Ultrack~\citep{bragantini_ultrack_2024} and revised manually. The cell division is indicated by a parent track splitting into two daughter tracks with the same parent track IDs. The last time-point of the parent track is considered the division event. The human annotator proofread and corrected the cell division events through visual inspection of the tracks in Ultrack GUI.

\section{Model architecture and training}
\label{sec:modelarch}

The model architecture has three main components: a spatial projection stem, an encoder backbone, and a multi-layer perceptron (MLP) head. The stem begins with a convolution layer with a kernel size of \( (5, 4, 4) \) for 3D datasets and \( (1, 4, 4) \) for 2D datasets and a stride of \( (5, 4, 4) \) for 3D datasets and \( (1, 4, 4) \) for 2D datasets, followed by a reshaping operation. This reshaping maps the down-sampled axial dimension to channels, efficiently projecting the anisotropic 3D input into a 2D feature map for encoding. The encoder backbone is adapted from the ConvNeXt Tiny architecture \citep{liu_convnet_2022}. The stem and head modules from ConvNeXt are removed, and the backbone outputs a 768-dimensional embedding vector \( \z \). The 768-dimensional vector \( \z \in \mathbb{R}^{768} \) is projected onto a lower 32-dimensional vector \( \p \in \mathbb{R}^{32} \) through a 2-layer MLP head, which helps speed up training \citep{chen_simple_2020}. To estimate the computational cost, we profiled the full DynaCLR encoder using the ptflops package \citep{sovrasov_ptflops_2024}. For input patches of size 2×15×256×256, the forward pass requires approximately 754 GFLOPs. Each training step, including the the gradient computations, is estimated at ~2.26 TFLOPs. The total training cost per model is approximately 0.5–1 PFLOPs for ~100K iterations.

We use the ImageNet \citep{huggingface_huggingfacepytorch-image-models_2024} and OpenPhenom-S \citep{kraus_masked_2023} pretrained models just for inference (\autoref{fig:imagenet-pretrained-ALFI}, \autoref{fig:imagenet-pretrained-infection}, \autoref{tab:app_sampling_strategies_alfi}
\autoref{tab:app_sampling_strategies_infection}.)

The models summarized in \autoref{tab:summary-of-models} were trained with a mini-batch size of 256, using the AdamW optimizer \citep{loshchilov_decoupled_2019}, and a learning rate of \( 2 \times 10^{-5} \). We used the HPC cluster on-premises using 2-4 GPUs with the distributed data parallel (DDP) strategy. A temperature of 0.3 was optimized for ALFI models to prevent overfitting with NT-Xent loss, as it is a small dataset. Other models with NT-Xent loss were trained with a temperature of 0.5. The margin of 0.5 was used in computational experiments that used the triplet loss. The time for model training depends on the size of the dataset, varying from an hour for the cell cycle model with the ALFI dataset to around 48 hours for the infection and organelle remodeling models. Due to the computational expense of training on the larger datasets we use the ALFI dataset for our time sampling experiments, which allows for significantly faster training times.

\subsection{Model library}
The paper reports multiple models (DynaCLR-*) trained with diverse channels from the above datasets, depending on the biological prediction task. We organize the models \autoref{tab:summary-of-models} based on the training and test data. The DynaCLR-DENV-* models were trained with a 30 minute sampled dataset and tested using 10 minute sampled data.

\begin{table}[H]
\caption{\textbf{Summary of models:} DynaCLR models organized by training and test data}
\label{tab:summary-of-models}
\centering
% \resizebox{\textwidth}{!}{%
\begin{tabular}{@{}llccl@{}}
\toprule
\textbf{Model name} & \textbf{Training data} & \textbf{Test data} & \textbf{Results shown in} \\
\midrule
DynaCLR-ALFI & U2OS (from ALFI) & HeLa + RPE1 (from ALFI) & Fig.~\ref{fig:contrastive-sampling}d, Fig.~\ref{fig:alfi-PHATE} \\
DynaCLR-microglia & Microglia (IL-17, IF-$\beta$) &  Microglia (glioblastoma) & Fig.~\ref{fig:contrastive-sampling}e \\
DynaCLR-DENV-VS+Ph & Phase+Viral Sensor & Phase+Viral sensor & Fig.~\ref{fig:generalization-infection} \\
DynaCLR-DENV-ER+Ph & Phase+SEC61 & Phase+SEC61 & Fig.~\ref{fig:organelle-dynamics}, \ref{fig:organelle-remodelling-phate} \\
DynaCLR-DENV-ER & SEC61 & SEC61 & Fig.~\ref{fig:alignment-embeddings-organelle-only} \\
DynaCLR-Teacher-VS & Viral Sensor & Viral Sensor & Fig.~\ref{fig:knowledge-distillation} \\
DynaCLR-Student-Ph & Phase & Phase & Fig.~\ref{fig:knowledge-distillation} \\
\bottomrule
\end{tabular}
% }
\end{table}

\begin{table}[H]
\caption{\textbf{Model input specifications:} The table provides a summary of input specifications for DynaCLR models. The input channels, patch size, and z-range for models are listed.}
\label{tab:model-input}
\centering
% \resizebox{\textwidth}{!}{%
\begin{tabular}{@{}lcccc@{}}
\toprule
\textbf{Model name} & \textbf{Input channels} & \textbf{Patch size (YX)} & \textbf{Z range} \\
\midrule
DynaCLR-ALFI & DIC & 128×128 & [0–1] \\
DynaCLR-microglia & Phase & 96×96 & [0–1] \\
DynaCLR-DENV-VS+Ph & Phase + sensor & 160×160 & [15–45] \\
DynaCLR-DENV-ER+Ph & Phase + SEC61 & 192×192 & [15–45] \\
DynaCLR-DENV-ER & SEC61 & 192×192 & [15–45] \\
DynaCLR-Teacher-VS & Viral Sensor & 160×160 & [15–45] \\
DynaCLR-Student-Ph & Phase & 160×160 & [15–45] \\
\bottomrule
\end{tabular}
% }
\end{table}

\subsection{Sampling and augmentation of patches of single cells}
\label{sec:sampling-appendix}

3D imaging volumes are cropped around the centroids of the tracking nodes to form single-cell patches.
We normalize the input image to reduce variability from experimental conditions. We rescale the viral sensor channel so that the median intensity is 0, and the 99th percentile intensity is 1. This normalization is more robust to extreme intensities in the fluorescence image, as well as variation in background fluorescence levels. The quantitative phase channel is normalized so that each field-of-view (FOV) has zero mean and unit standard deviation. The phase image is already normalized during reconstruction \citep{guo_revealing_2020}, and this extra standardization step ensures proper input numerical range for the model.
We use a larger initial crop to ensure no padding is included in the final input patch after spatial augmentations.
We apply extensive augmentations (\autoref{tab:augmentations})
at training time to simulate variations induced by the imaging system
and other non-biological conditions.
The input patch size after augmentations is optimized for reducing the influence from background and neighboring cells while focusing on the peri-nuclear region of the cell,
where the majority of infection-related changes, such as viral sensor re-localization and ER remodeling, are captured. The patch sizes and z-ranges used for different models are listed in \autoref{tab:model-input}.

\begin{table}[h]
\caption{
    Augmentations applied to image patches.
    Parameters are supplied to respective MONAI \citep{cardoso_monai_2022} transforms,
    where $\alpha$ denotes scaling factor,
    $\theta$ denotes rotation (radians),
    $s$ denotes shearing,
    $\gamma$ denotes gamma value,
    $\sigma$ denotes the standard deviation of the Gaussian distribution,
    and $p$ denotes the probability of applying the random transform.
}
\label{tab:augmentations}
\centering
\begin{tabular}{@{}ll@{}}
\toprule
\textbf{Augmentation Type} & \textbf{Parameters} \\
\midrule
Random spatial scaling     & $\alpha_x, \alpha_y \in [-0.3, 0.3]$, $p=0.8$ \\
Random rotation            & $\theta_z \in [0, \pi]$, $p=0.8$ \\
Random shearing            & $s_x, s_y \in [0, 0.01]$, $p=1.0$ \\
Random contrast adjustment & $\gamma \in [0.8, 1.2]$, $p=0.5$ \\
Random intensity scaling   & $\alpha \in [-0.5, 0.5]$, \\
                           & $p_\text{Phase}=0.5$, $p_\text{Sensor}=0.7$ \\
Gaussian smoothing         & $\sigma_x, \sigma_y \in [0.25, 0.75]$, $p=1.0$ \\
Gaussian noise addition    & $\sigma_\text{Phase} \in [0, 0.2]$, \\
                           & $\sigma_\text{Sensor} \in [0, 0.5]$, $p=0.5$ \\
\bottomrule
\end{tabular}
\end{table}

\clearpage

\section{Metrics}
\label{sec:metrics}
To characterize the temporal continuity and variability of a tracked cell \( i \) in the embedding space, we analyze its trajectory \( \z_i(t) \in \mathbb{R}^d \) via cosine distance. We measure the distance between the embeddings of two cells via cosine distance:
\begin{equation}
D_{ij}(t_a, t_b) = 1 - \frac{ \z_i(t_a) \cdot \z_j(t_b) }
{ \left\| \z_i(t_a) \right\| \cdot \left\| \z_j(t_b) \right\| },
\quad \text{for } t = 1, \ldots, T{-}1.
\label{eq:cosine_displacement}
\end{equation}

We assess the effect of the contrastive sampling method and the loss functions on the temporally regularized embedding space using the pairwise cosine distance between random and adjacent timepoints $t$ with the following metrics:

\paragraph{Smoothness:}
Smoothness quantifies how much short-term variation exists relative to overall variation in the embedding space. We compute the ratio of the mean distance between adjacent timepoints in each trajectory to the mean distance between randomly sampled timepoints from the same trajectory:
\begin{equation}
\text{Smoothness} = \frac{D_{adj}}{D_{rand}}= \frac{ \mathrm{mean}_{i,\, t} \left[ D_i(t, t{+}1) \right] }
{ \mathrm{mean}_{i,\, (t_a, t_b)} \left[ D_{i}(t_a, t_b) \right] }
\label{eq:sensitivity}
\end{equation}
where \( D_i(t_a, t_b) \) is the cosine distance between embeddings at a randomly sampled pair of embeddings at timepoints \( t_a \) and \( t_b \), and \( t_a \ne t_b \). A lower value indicates a temporally smooth embedding space.

\paragraph{Dynamic Range ($DR$):}
The dynamic range quantifies how much variation is captured in the embedding space over time. 
It is defined as the difference between the peaks of the embedding distance distributions for randomly selected frame pairs and adjacent frame pairs (\autoref{fig:alfi-cosine}), computed over all tracks in the datasets. The peaks were identified using Gaussian KDE.

% \paragraph{Diffusion rate ($\alpha$):}
% We assess the temporal smoothness of the embeddings by computing the mean squared displacement (MSD) in the embedding space. The intuition for this metric is that the cells are undergoing directed change in the presence of perturbation and this metric characterizes how fast the change is represented by the embeddings. We use Euclidean distance measure and compute the slope of the log-log mean squared displacement (MSD) curve:
% \begin{equation}
%     \text{MSD}(\Delta t) = \frac{1}{N} \sum_{i=1}^{N} \left\| \z_i(t{+}\Delta t) - \z_i(t) \right\|^2,
%     \quad \text{then} \quad
%     \alpha = \frac{d \log \text{MSD}}{d \log \Delta t}
% \label{eq:MSD_smoothness}
% \end{equation}

% We compute the slope \( \alpha \) over short time intervals to ensure stability, since MSD estimates become noisy at large \(\Delta t\). Lower values of \( \alpha \) indicate more temporally regular embeddings. Higher slope implies that rate of change of embeddings is high on a shorter time scale.

\section{Dynamic Time Warping (DTW)}
\label{sec:dtw}
DTW is particularly well-suited for capturing heterogeneous cellular responses to perturbations, where variations in progression speed or trajectory length are common despite underlying conserved patterns. It computes an optimal, non-linear alignment path between two sequences by minimizing the cumulative pairwise distance, while allowing for local stretching and compression along the pseudo-time axis~\citep{sakoe_dynamic_1978}. 

Each tracked cell \( i \) has a sequence of embeddings:
\[
\mathbf{Z}_i = \left( \z_i(1), \z_i(2), \ldots, \z_i(T_n) \right), \quad \z_i(t) \in \mathbb{R}^d,
\]
where \( d \) is the embedding dimension and \( T_n \) is the number of timepoints in the trajectory.

We slide a window of length \( m \) (equal to the length of the reference trajectory \( \mathbf{Z}_{\text{ref}} \)) across each \( \mathbf{Z}_i \) and compute the DTW alignment cost at each offset:
\[
\text{Cost}_i(s) = \text{DTW} \left( \mathbf{Z}_{\text{ref}},\; (\z_i(s), \ldots, \z_i(s{+}m{-}1)) \right),
\quad \text{for } s = 1, \ldots, T_i - m + 1.
\]

We identify the window with the lowest alignment cost:
\[
s_i^* = \arg\min_s \text{Cost}_i(s),
\]
and extract the best-matching sequence of embeddings:
\[
\mathbf{Z}_i^{*} = (\z_i(s_i^*), \ldots, \z_i(s_i^*{+}m{-}1)).
\]

This sequence is then aligned to \( \mathbf{Z}_{\text{ref}} \) using the DTW alignment path, yielding a warped trajectory \( \mathbf{Z}_i^{\text{aligned}} \) in the reference pseudo-time. This allows asynchronous dynamics across single cells to be mapped into a shared temporal coordinate system.

\begin{table}[h]
\small
\caption{\textbf{Performance of DynaCLR-ALFI models:} F1 score of linear classification, pairwise distance of adjacent and random frames, dynamic range, and smoothness (as defined in Appendix \autoref{sec:metrics}) for interphase vs. mitosis classification for different time samplings, different losses, and sampling strategies}
\label{tab:app_sampling_strategies_alfi}
\begin{center}
\begin{tabular}{p{4.5cm}|c|c|c}
\multicolumn{1}{|p{4.5cm}|}{\bf Experiments}  & 
\multicolumn{1}{|p{2cm}|}{\bf F1 score $\uparrow$}  &
\multicolumn{1}{|p{2cm}|}{\bf Smoothness $\downarrow$} &
\multicolumn{1}{|p{2cm}|}{\bf DR $\uparrow$} 
\\ \hline \\
Time Aware + NT-Xent (\(\tau = 7\)) & 98.09 & \textbf{0.12} & 1.29 \\
Time Aware + NT-Xent (\(\tau = 14\)) & 97.31 & 0.13 & 1.18 \\
Time Aware + NT-Xent (\(\tau = 28\)) & \textbf{98.84} & 0.14 & 1.1 \\
Time Aware + NT-Xent (\(\tau = 56\)) & 97.82 & 0.15 & 1.17 \\
Time Aware + NT-Xent (\(\tau = 91\)) & 98.43 & 0.15 & 1.07 \\
Time Aware + triplet (\(\tau = 7\)) & 98.26 & 0.15 & 0.95 \\
Time Aware + triplet (\(\tau = 14\)) & 98.26 & 0.15 & 0.86 \\
Time Aware + triplet (\(\tau = 28\)) & 98.09 & 0.18 & 0.94 \\
Time Aware + triplet (\(\tau = 56\)) & 98.52 & 0.17 & 1.03 \\
Time Aware + triplet (\(\tau = 91\)) & 98.26 & 0.16 & 0.85 \\
Cell aware + triplet & 97.74 & 0.13 & \textbf{1.38} \\
Classical + NT-Xent (no tracking) & 97.57 & \textbf{0.12} & 1.27 \\
Classical + triplet (no tracking) & 98.34 & 0.13 & 0.72 \\
ImageNet pretrained & 98.48 & 0.35 & 0.8 \\
OpenPhenom-S/16 pretrained & 97.5 & 0.27 & 0.92 \\
\end{tabular}
\end{center}
\end{table}

\begin{table}[h]
\small
\caption{\textbf{Performance of DynaCLR-DENV-VS+Ph models:} F1 score of linear classification of pairwise distance of adjacent and random frames, dynamic range, and smoothness for infection state classification for different losses, and sampling strategies. For OpenPhenome and ImageNet,  only the viral sensor channel was used as input.}
\label{tab:app_sampling_strategies_infection}
\begin{center}
\begin{tabular}{p{3cm}|c|c|c|c|c|c}
\multicolumn{1}{|p{2cm}|}{\bf Experiments}  & 
\multicolumn{1}{|p{2.5cm}|}{\bf F1 score $\uparrow$}  &
\multicolumn{1}{|p{2cm}|}{\bf Smoothness $\downarrow$ }&
\multicolumn{1}{|p{2cm}|}{\bf DR $\uparrow$} 
\\ \hline \\
Time Aware + NT-Xent & \textbf{98.40} &  \textbf{0.15} & \textbf{1.33}  \\
Time Aware + triplet & 96.56 & 0.16 & 1.07  \\
Cell aware + triplet & 98.24 & 0.24 & 1.07  \\
Classical + NT-Xent & \textbf{98.41} &  0.32 & 1.23  \\
Classical + triplet & 98.07 &  0.23 & 1.03  \\
ImageNet pretrained & 97.82 & 0.47 & 0.74  \\
OpenPhenome-S/16 & 95.2 & 0.32 & 1.18  \\
Supervised semantic segmentation model & 83 & - & - \\
\end{tabular}
\end{center}
\end{table}

\section{Software Demo}
\label{sec:additional_resources}
We provide a software demonstration via an anonymous Google Drive link, because the test dataset and model checkpoints are >30GB in size. Please see README.md in:  
\url{https://drive.google.com/drive/folders/1SeQcWQcTF3Xfvz4XU_2DzMzGSxc-Hkgb}

\section{Supplementary Figures}

\begin{AppendixFig}[H]
    \centering
    \includegraphics[width=\linewidth]{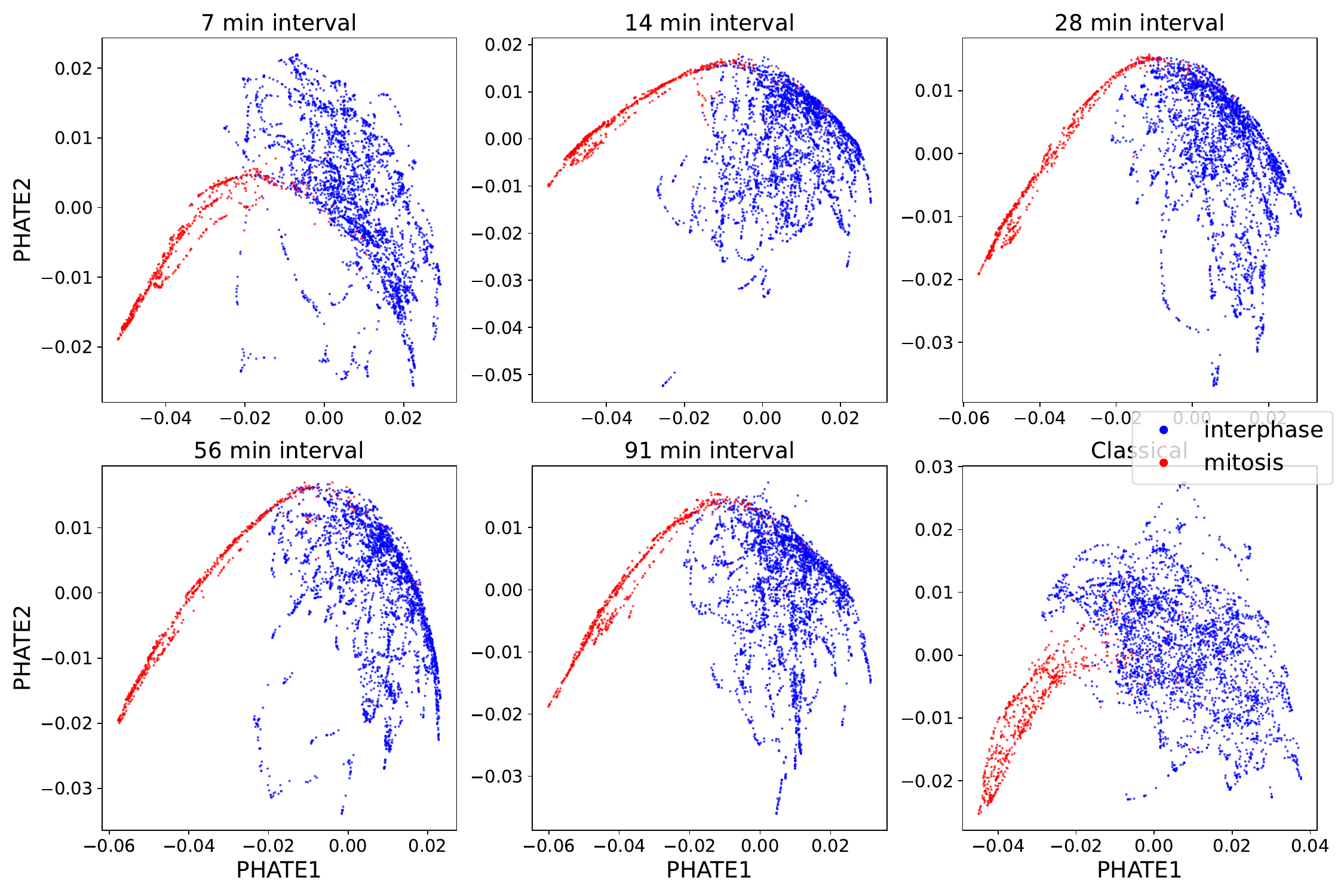}
    \caption{
        \textbf{PHATE embeddings of cells under different sampling strategies.} 
        These panels show the PHATE embeddings for cells across various time shifts (\( \tau \)) and sampling strategies. Time-aware sampling (e.g., 7 min, 21 min, 28 min intervals) results in embeddings where similar cell states are closer and more continuous in embedding space, potentially reflecting improved temporal alignment. In contrast, classical sampling exhibits greater scattering and discontinuity in cell state trajectories. The embedding continuity highlights the ability of time-aware sampling to better preserve temporal relationships between cell states compared to classical methods.
    }
    \label{fig:alfi-PHATE}
\end{AppendixFig}

\begin{AppendixFig}[H]
    \centering
    \includegraphics[width=\linewidth]{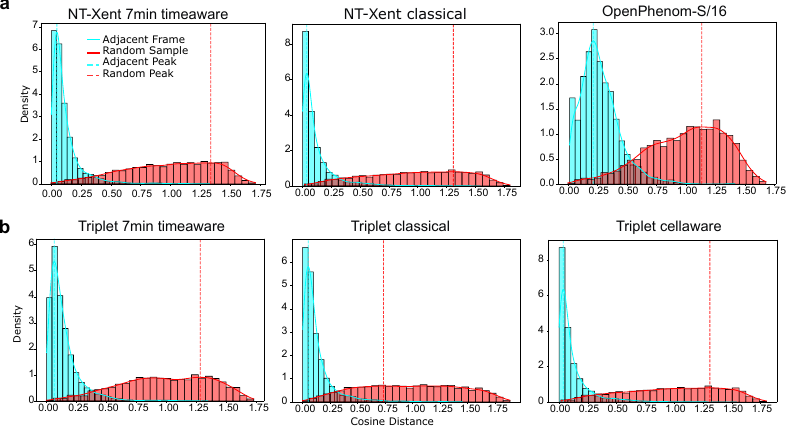}
\caption{
\textbf{Distribution of cosine distance between adjacent and random frames for various sampling strategies used on ALFI dataset.}
(a) Distribution of cosine distances for time-aware and classical models with NT-Xent loss, with peaks of distributions indicated by dashed lines. 
(b) Distributions of cosine distance for time-aware, classical, and cell-aware models with triplet loss, with peaks of the distribution indicated by dashed lines. The dynamic range of the embedding is the distance between the peaks of the distributions of random and adjacent frames.}
\label{fig:alfi-cosine}
\end{AppendixFig}

\begin{AppendixFig}[H]
    \centering
    \includegraphics[width=\linewidth]{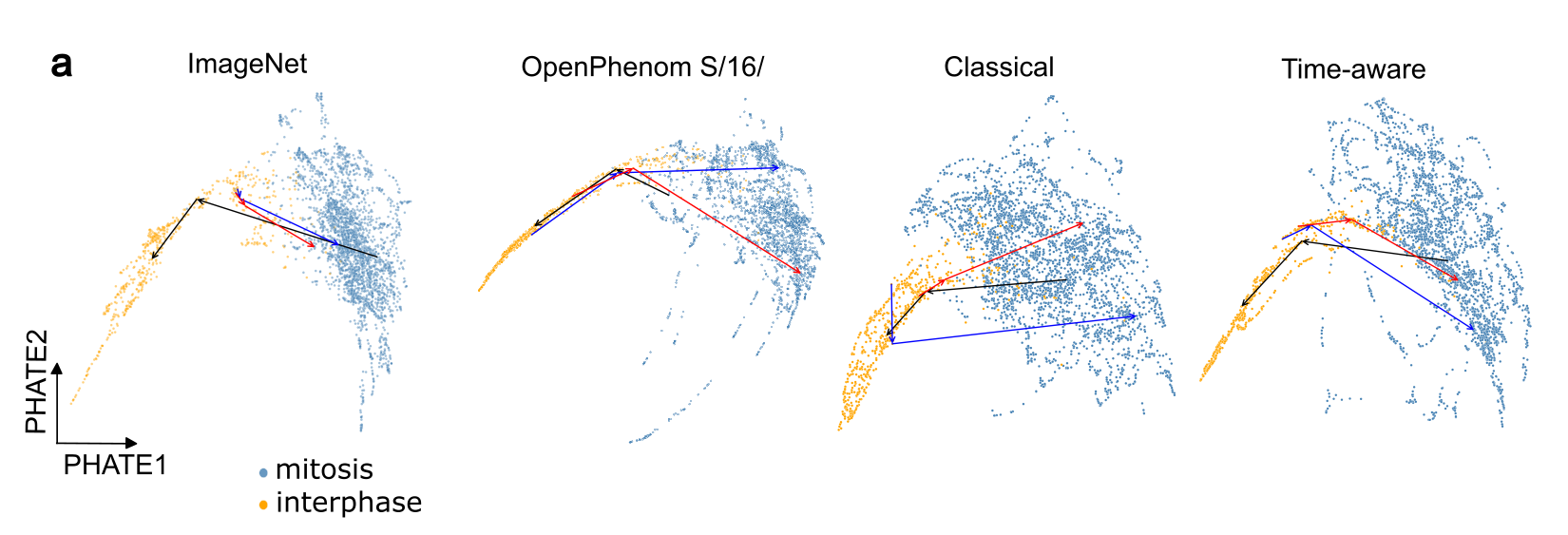}
    \caption{
        \textbf{The ImageNet pre-trained model performs comparably to the classical model
in classifying infection cell division}(a) PHATE maps display distinct clustering of cells in interphase and mitosis, as indicated by color-coded human annotations. 
    }
    \label{fig:imagenet-pretrained-ALFI}
\end{AppendixFig}

% \begin{AppendixFig}[H]
%     \centering
%     \includegraphics[width=\linewidth]{fig/appendix_compare_timesampling.pdf}
% \caption{
% \textbf{Evaluation of embedding displacement and temporal smoothness across sampling strategies.}
% (a) Mean squared displacement (MSD) of embedding over increasing time shifts (\(\Delta t\)) for various time intervals of time-aware sampling strategy and classical sampling strategy. 
% (b) MSD vs time shifts of different sampling strategies in a) on log scale.
% (c) slope at early and mid-section of time shifts of log of MSD vs time for the different time intervals and sampling strategies.}
%     \label{fig:alfi-displacements}
% \end{AppendixFig}

\begin{AppendixFig}[H]
    \centering
    \includegraphics[width=0.9\linewidth]{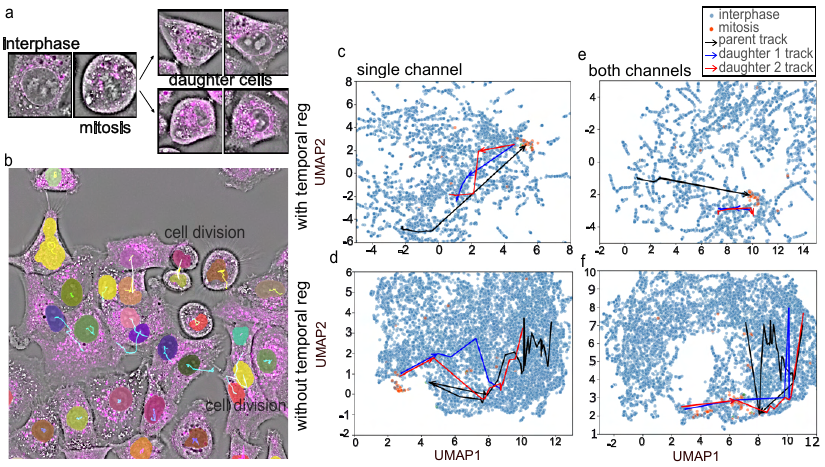}
    \caption{\textbf{Detection of rare events, e.g., cell division}:(a) The morphology of the cell changes over time during the transition between interphase and mitosis. (b) Ultrack tracks the cell over time and captures mitosis. White tracks indicate cell divisions. (c--f) The trajectory of one parent cell (black track) dividing into two daughter cells (blue and red tracks) overlaid on the UMAP from models using phase channel and a combination of phase and viral sensor channels, and with and without temporal regularization, illustrates that temporal regularization leads to smooth trajectories and better clustering with just the phase channel. However, we cannot classify the cell division event here due to the very low number of mitosis events in the dataset.}
     % \textbf{TODO: SP}
    \label{fig:cell-division-a549}
\end{AppendixFig}

\begin{AppendixFig}[H]
    \centering
    \includegraphics[width=\linewidth]{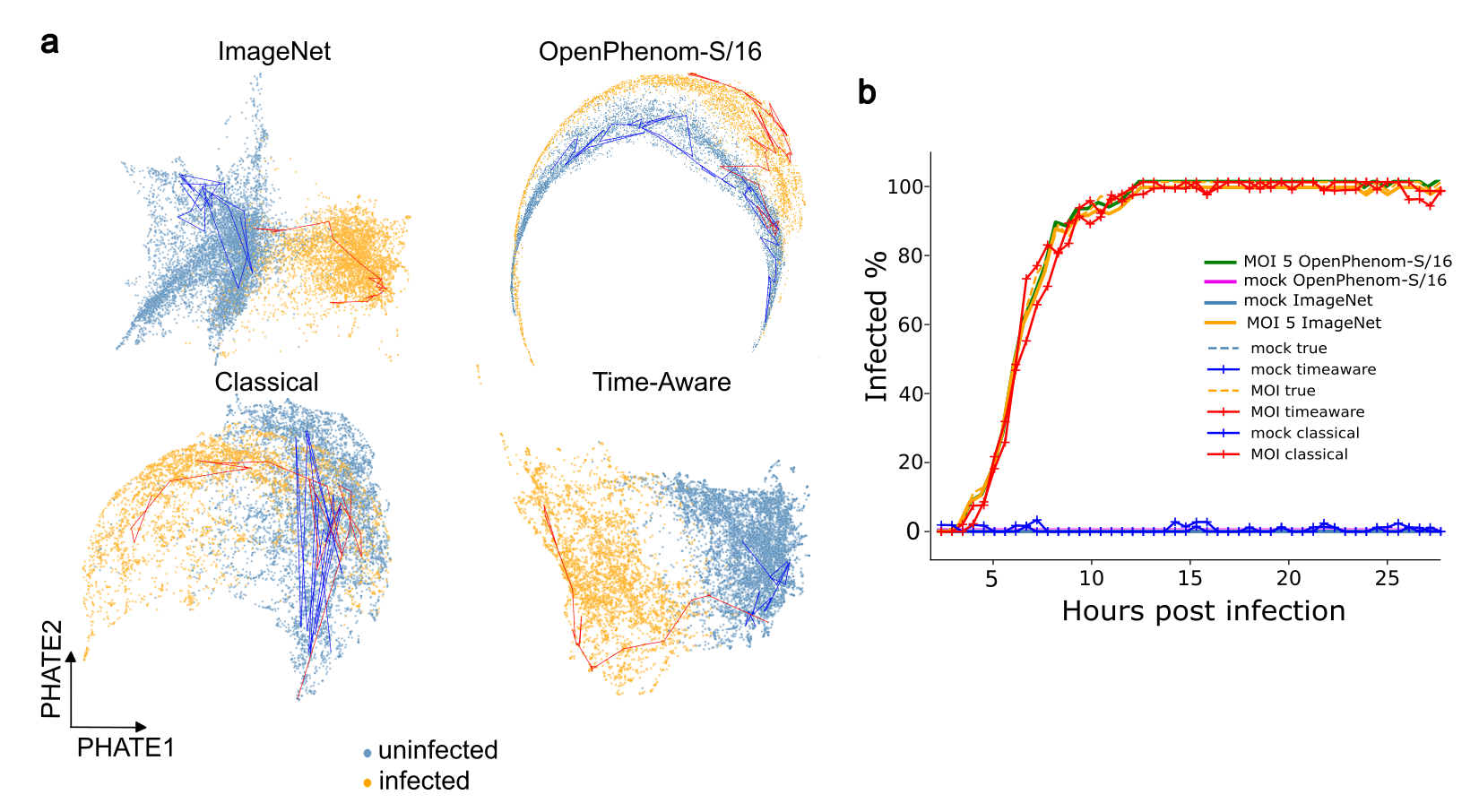}
    \caption{
        \textbf{DynaCLR classical and time-aware models perform comparably to  pre-trained at infection state classification}
        a) PHATE embeddings reveals clear clustering of infected and uninfected cells based on features from the ImageNet pre-trained model, OpenPhenom-S/16 pre-trained model, DynaCLR Classical and Time-aware models using a linear classifier trained with sparse annotations. b) Infection percentage over time shows an exponential increase, consistent with expected infection dynamics. OpenPhenom and ImageNet models take only the viral sensor channel as input.
    }
    \label{fig:imagenet-pretrained-infection}
\end{AppendixFig}

\begin{AppendixFig}[H]
    \centering
    \includegraphics[width=\linewidth]{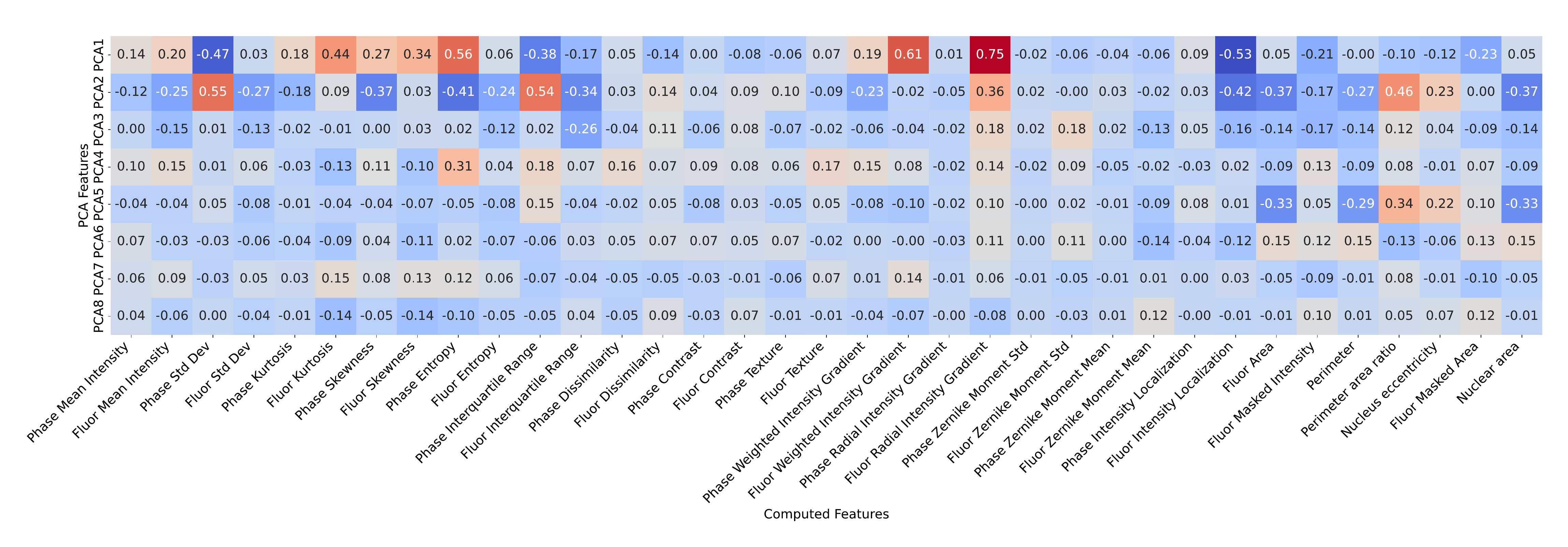}
    \caption{
        \textbf{Principal components vs computed features for the viral sensor and phase model}:
        Principal components correlate with interpretable image features such as radial intensity profile, area of fluorescence, and phase texture statistics, suggesting that the model captures biologically relevant variation.
    }
    \label{fig:PC_vs_CF_sensor}
\end{AppendixFig}

\begin{AppendixFig}[H]
    \centering
    \includegraphics[width=\textwidth]{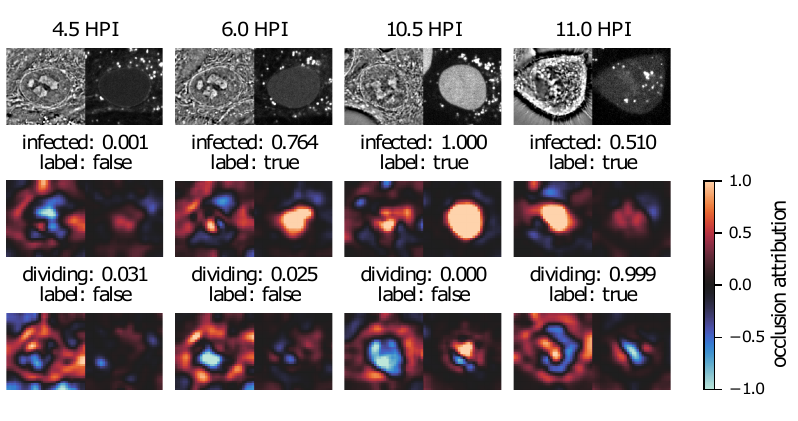}
    \caption{
        \textbf{Explanation of embeddings learned by DynaCLR:} Occlusion attribution of a cell undergoing infection and division:
        The first row shows a center slice of the input images at different time points,
        the second row shows attribution with an infection classification head,
        and the third row shows attribution with a division classification head.
        The titles show the predicted probability and true class.
    }
    \label{fig:explain-embeddings}
\end{AppendixFig}

\begin{AppendixFig}[H]
    \centering
    \includegraphics[width=\linewidth]{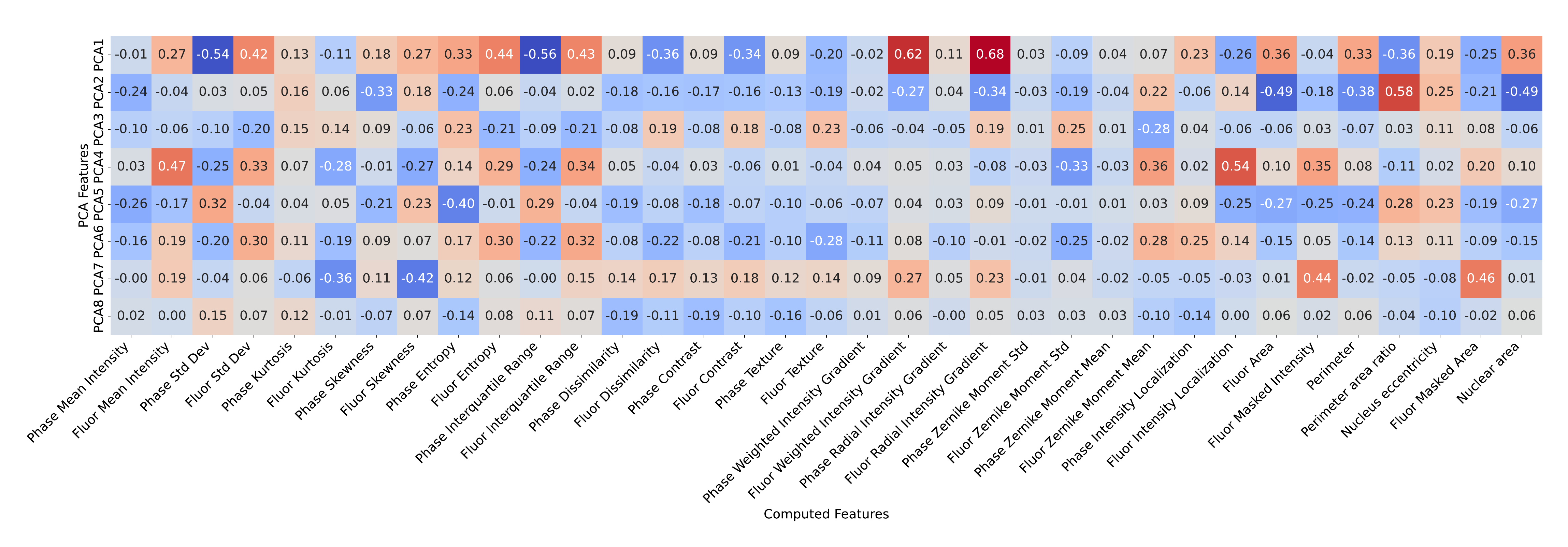}
    \caption{
        \textbf{Principal components vs computed features for the SEC61 and phase model}: Extracted a set of interpretable features (e.g. area, eccentricity, texture, mean intensity, etc.) from the phase and organelle channel and computed Spearman correlations between these features and the top 8 principal components of the learned embeddings.
    }
    \label{fig:PC_vs_CF_organelle}
\end{AppendixFig}

\begin{AppendixFig}[H]
    \centering
    \includegraphics[width=\textwidth]{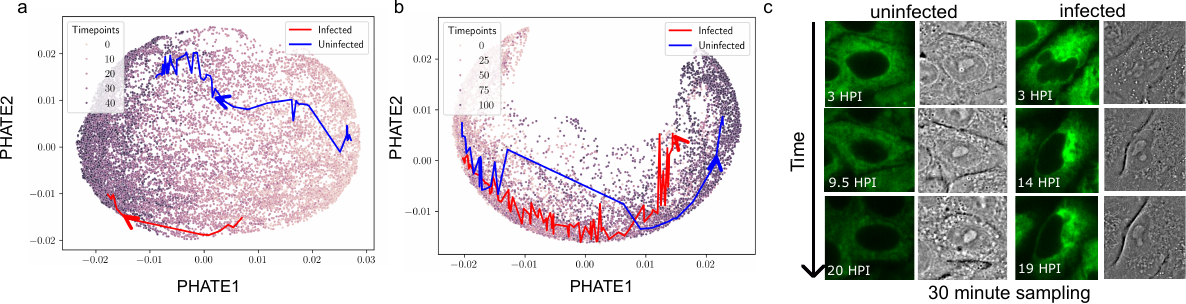}
    \caption{
        \textbf{Discovering ER remodeling due to infection:} Evaluation of the SEC61, endoplasmic reticulum (ER) marker, and quantitative phase model (a) in-distribution and (b) out-of-distribution. Pairs of SEC61 marker (green) and quantitative phase (gray) of uninfected and infected cells over time, showing how infection remodels and condenses the ER. 
    }
    \label{fig:organelle-remodelling-phate}
\end{AppendixFig}

\begin{AppendixFig}[H]
    \centering
    \includegraphics[width=\linewidth]{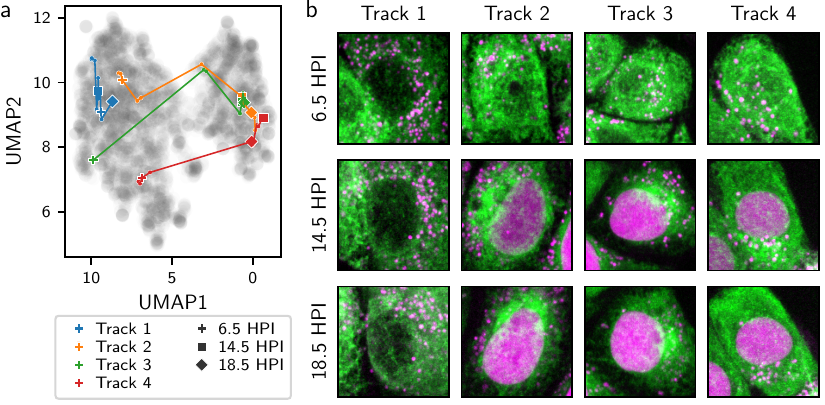}
    \caption{
        \textbf{Learned representation of the phase and viral sensor channels
        help exploration of organelle remodeling during infection.}
        (a) UMAP of learned features computed for
        mock and Dengue infected cells in the independent test dataset
        where the ER of cells is labeled with a fluorescent protein (SEC61-GFP).
        Track 1 from the mock well and tracks 2-4 from the Dengue infected well are highlighted.
        Cells other than the example tracks are marked in gray.
        (b) Snapshots from example tracks in (a),
        showing max-intensity projection of ER (green) and the viral sensor (magenta).
        In some of the infected cells (tracks 2 and 3),
        ER forms transient condensation.
    }
    \label{fig:organelle-dynamics-umap}
\end{AppendixFig}

\begin{AppendixFig}[H]
    \centering
    \includegraphics[width=\textwidth]{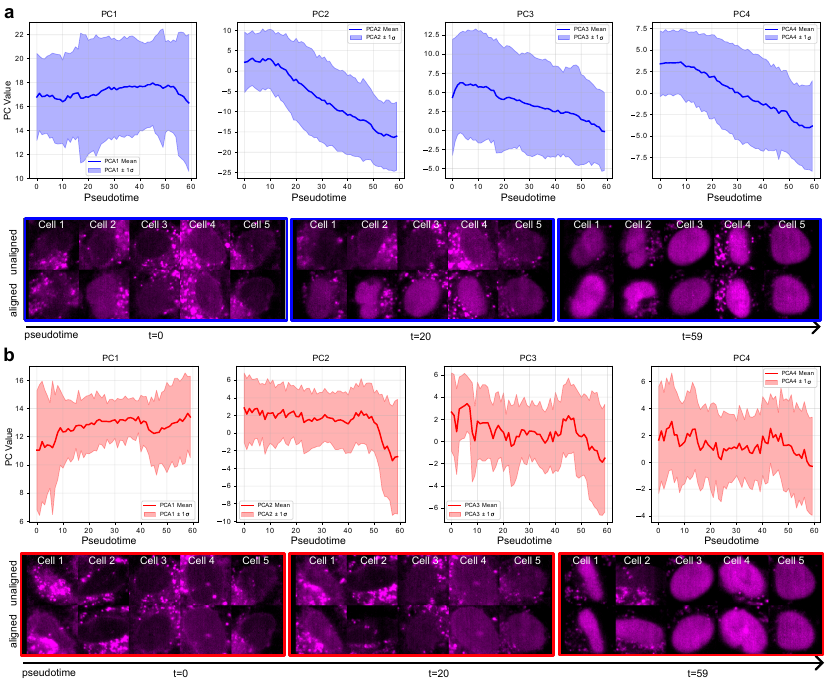}
    \caption{
    \textbf{Embedding alignment using viral sensor and phase channels.}  
    (a) Principal components of embeddings from the DynaCLR-DENV-VS+Ph model on the test dataset, aligned using Dynamic Time Warping (DTW). Embeddings were selected from sequences where infection was observed. The plots show the mean principal component value per timepoint (pseudo-time), with shaded bands representing one standard deviation. The top five aligned cells are visualized at pseudo-time points 0, 20, and 59, demonstrating consistent alignment across the timeline, including at t = 20.
    (b) Comparison with aligned principal components from OpenPhenom S/16. The top five aligned cells are shown, revealing difficulty in achieving alignment at pseudo-time t = 20.
    }
    \label{fig:alignment-model-comparison}
\end{AppendixFig}

\newpage

\begin{AppendixFig}[H]
    \centering
    \includegraphics[width=0.9\linewidth]{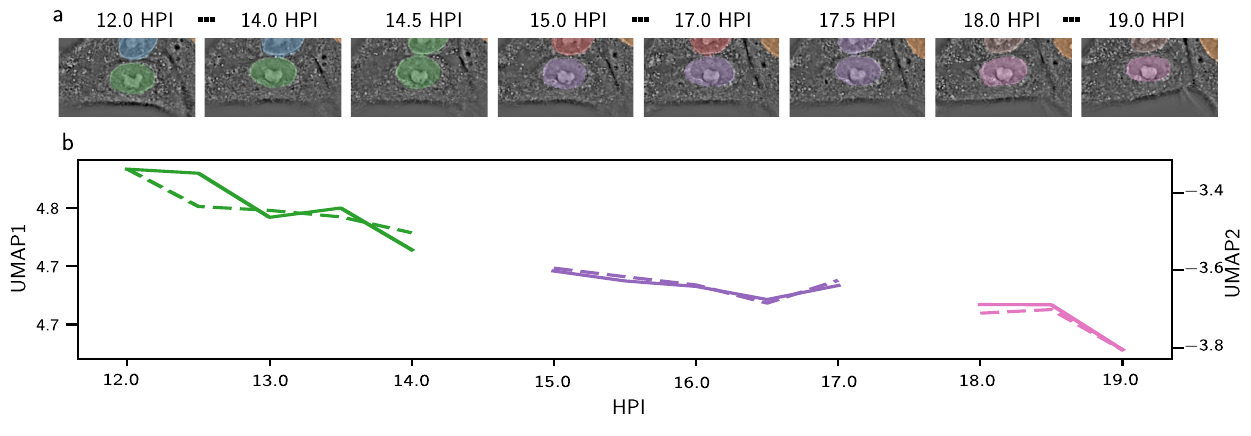}
    \caption{
        \textbf{DynaCLR embeddings are smooth even when tracking is erroneous:}
        (a) snapshots of a cell and its tracking labels over time.
        Note that the false fusion in 14.5 and 17.5 HPI frames
        caused subsequent false division and identity jump of the cell.
        (b) UMAP components 1 (solid line) and 2 (dashed line)
        over time for the falsely assigned tracks.
        The gaps correspond to false fusion events
        which shifts the centroid of the track towards the edge of the FOV,
        resulting in invalid patches.
        The UMAP components are smoothly transitioning over time,
        even though they are assigned to different tracks.
        }
    \label{fig:bridge-tracks}
\end{AppendixFig}

\begin{AppendixFig}[H]
    \centering
    \includegraphics[width=\textwidth]{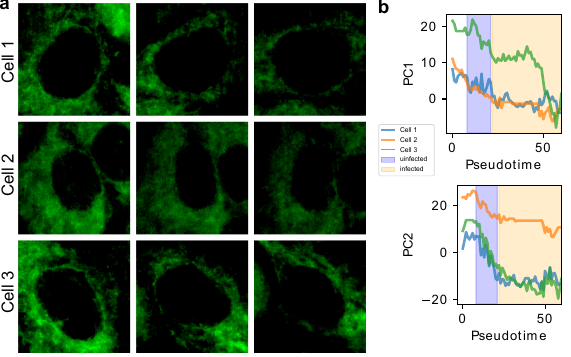}
    \caption{
        \textbf{Alignment of embeddings: Organelle model} Pseudo-time alignment of the SEC61 only model aligned based on the infection state.
    }
    \label{fig:alignment-embeddings-organelle-only}
\end{AppendixFig}

\pagebreak

\section{Videos}
\label{sec:videos}

\begin{Video}[H]
    \centering
    \includegraphics[width=\textwidth]
    {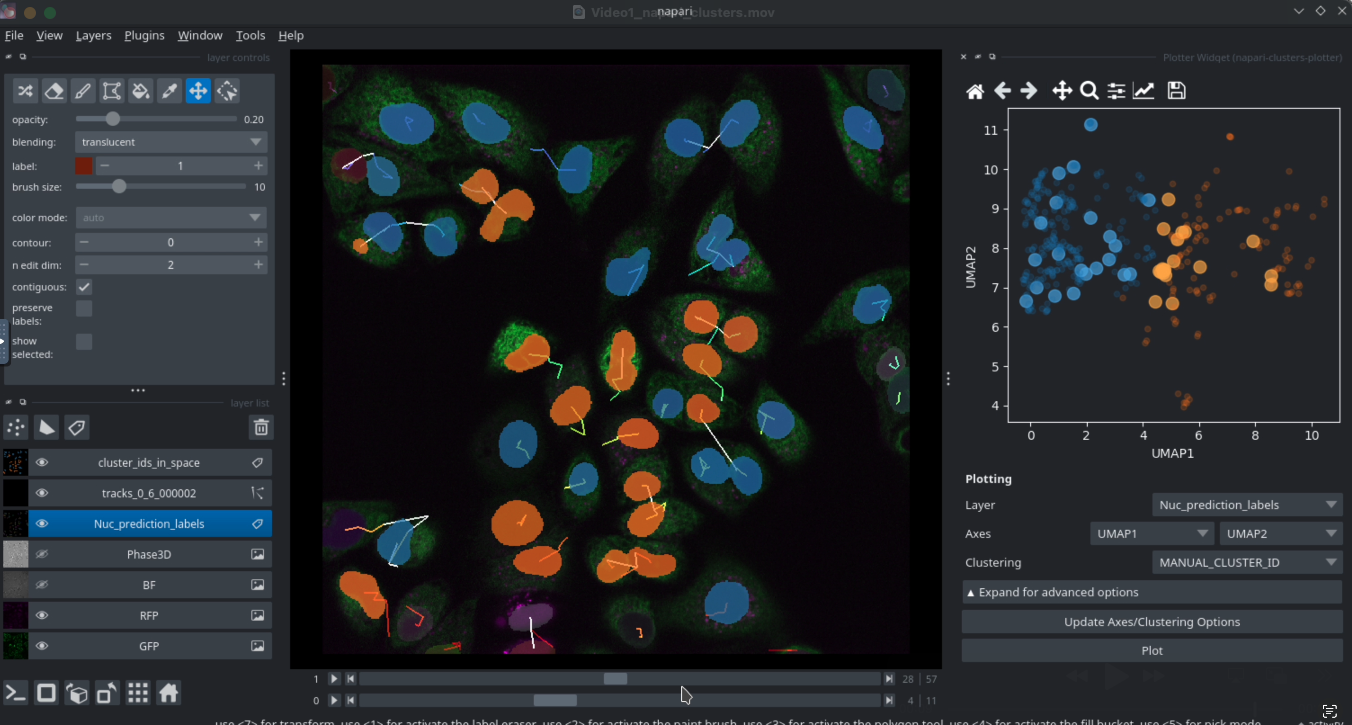}
    \caption{\textbf{Annotations with DynaCLR embeddings}:A napari workflow for interactive exploration and annotations. We developed a napari plugin to load images, tracks, and learned features. Then napari-clusters-plotter is used to interactively visualize the cell dynamics in latent space with reference to the morphological changes in real space.
    }
    \label{vid:napari}
\end{Video}

\begin{Video}[H]
    \centering
    \includegraphics[width=\linewidth]{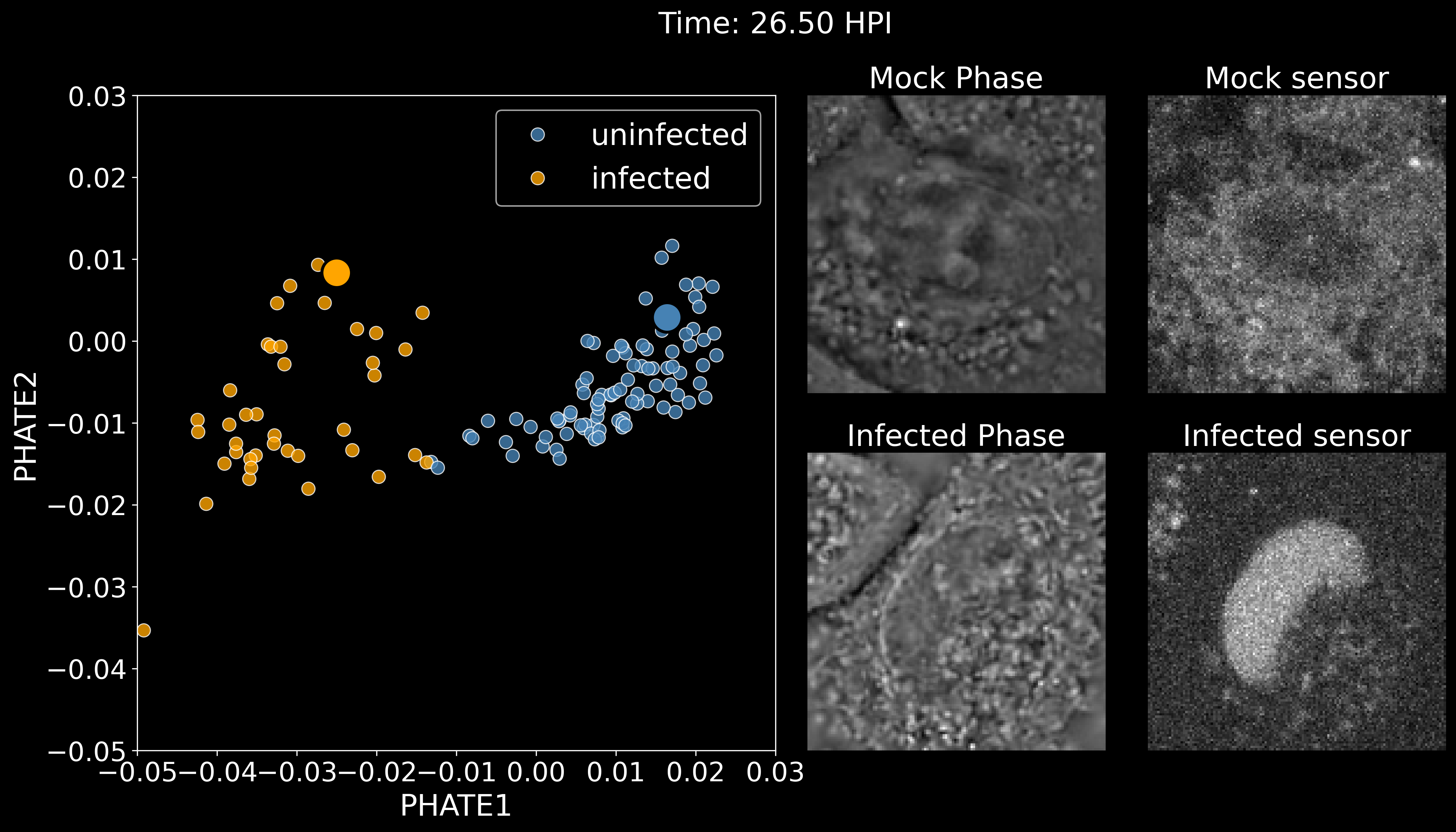}
    \caption{\textbf{Viral sensor dynamics}: Evolving dynamics of infection in unseen test data with time from a different microscope, colored by model prediction. Images show representative cells from mock and MOI 5 infected conditions.
    }
    \label{vid:viral-sensor}
\end{Video}

\begin{Video}[H]
    \centering
    \includegraphics[width=\linewidth]{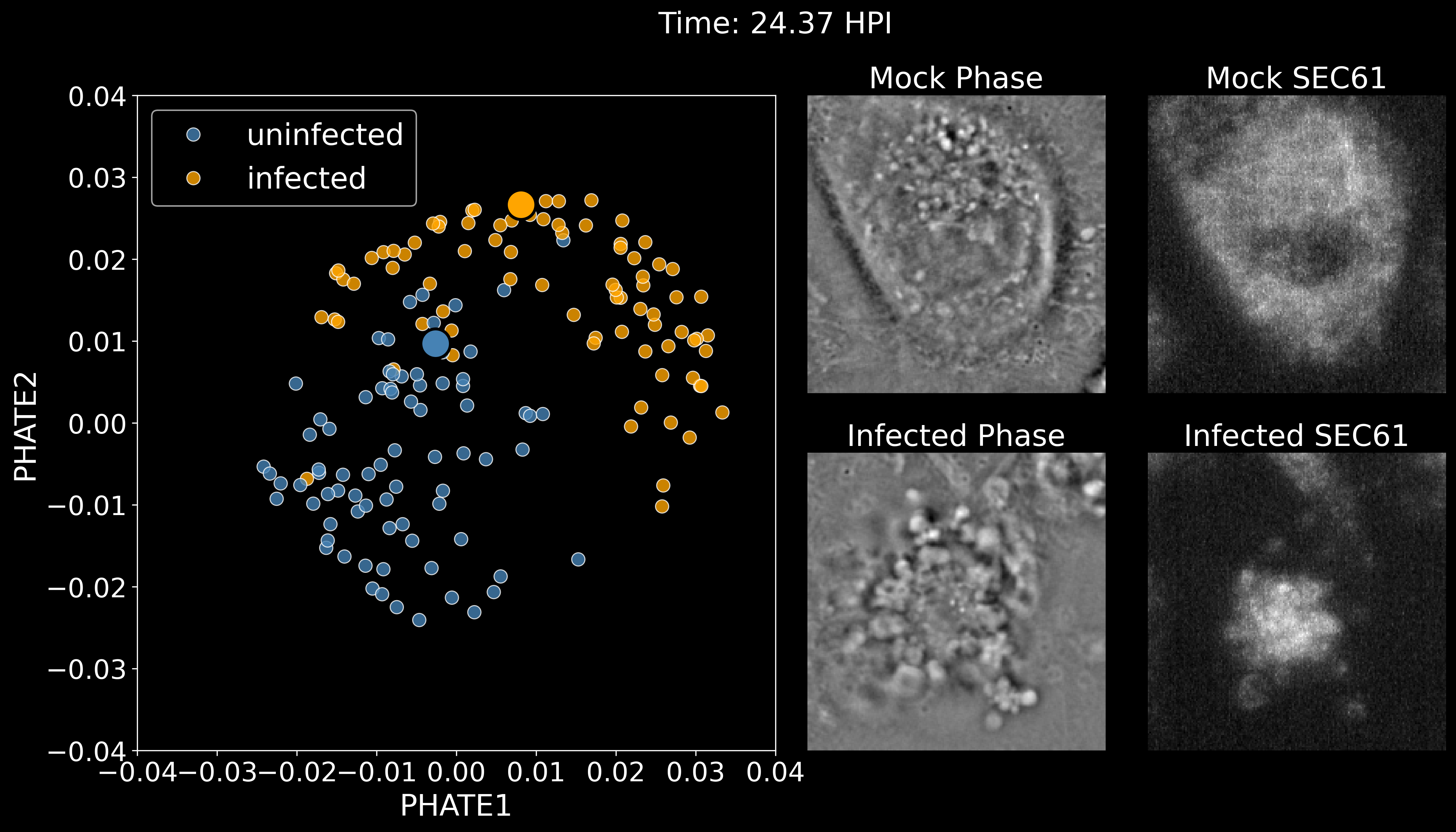}
    \caption{\textbf{Organelle remodeling}: Evolving dynamics of organelle architecture in 10-minute resolved test data with time, colored by model prediction of infection. Images show a representative cell from the mock condition, condensing at the end of the movie for division, and another cell from MOI 5 infected conditions with ER remodeling due to infection, eventually leading to cell death.
    }
    \label{vid:organelle-remodelling}
\end{Video}

\begin{Video}[H]
    \centering
    \includegraphics[width=\textwidth]{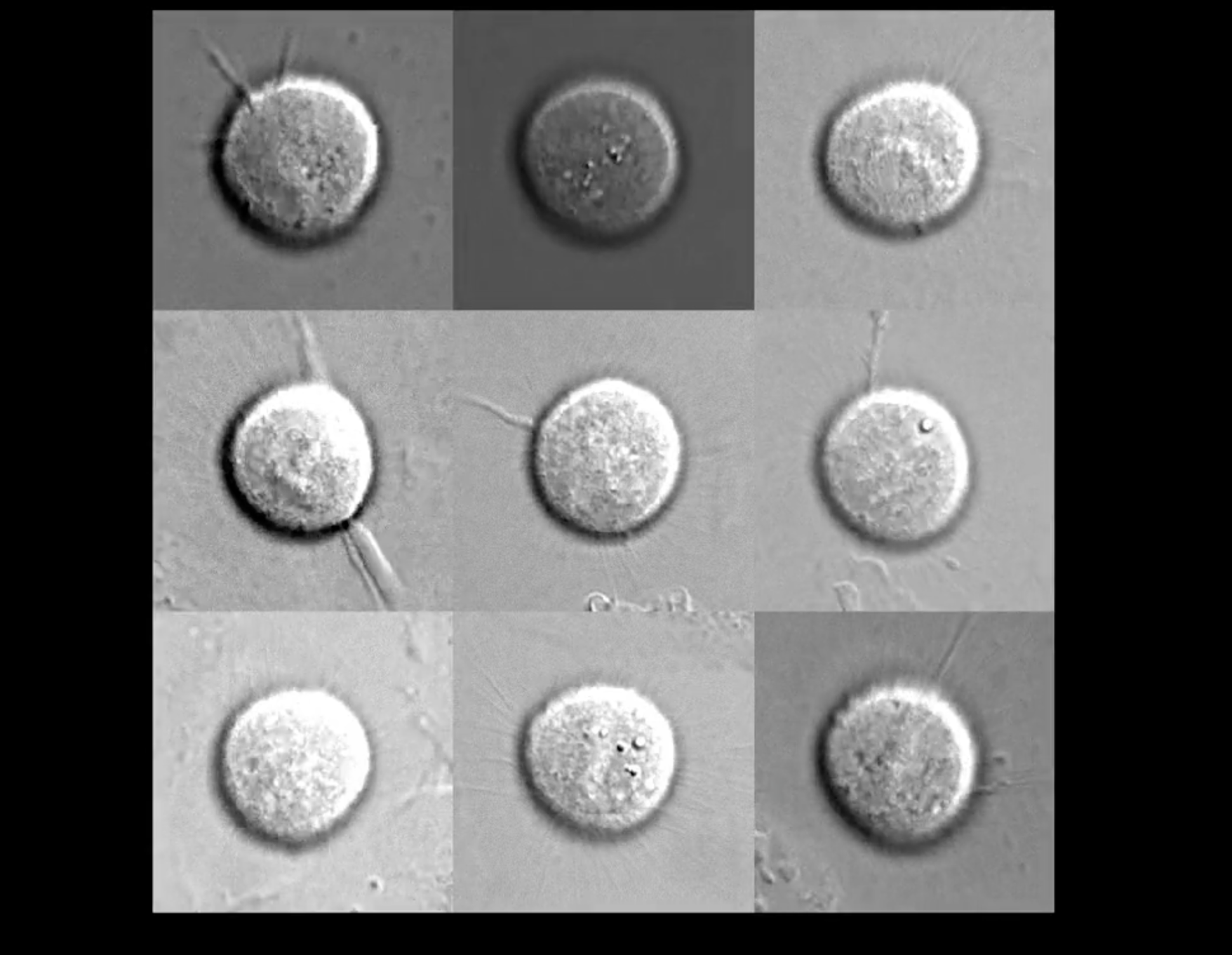}
    \caption{\textbf{Digital synchronization of cell-division}: Cells from the ALFI dataset are digitally synchronized using dynamic time warping (DTW) to align their trajectories across the mitotic transition. Each cell is aligned to a reference trajectory (bottom right), highlighting consistent morphological progression between interphase and mitosis.
    }
    \label{vid:dtw-alfi}
\end{Video}

\begin{Video}[H]
\includegraphics[width=\textwidth]{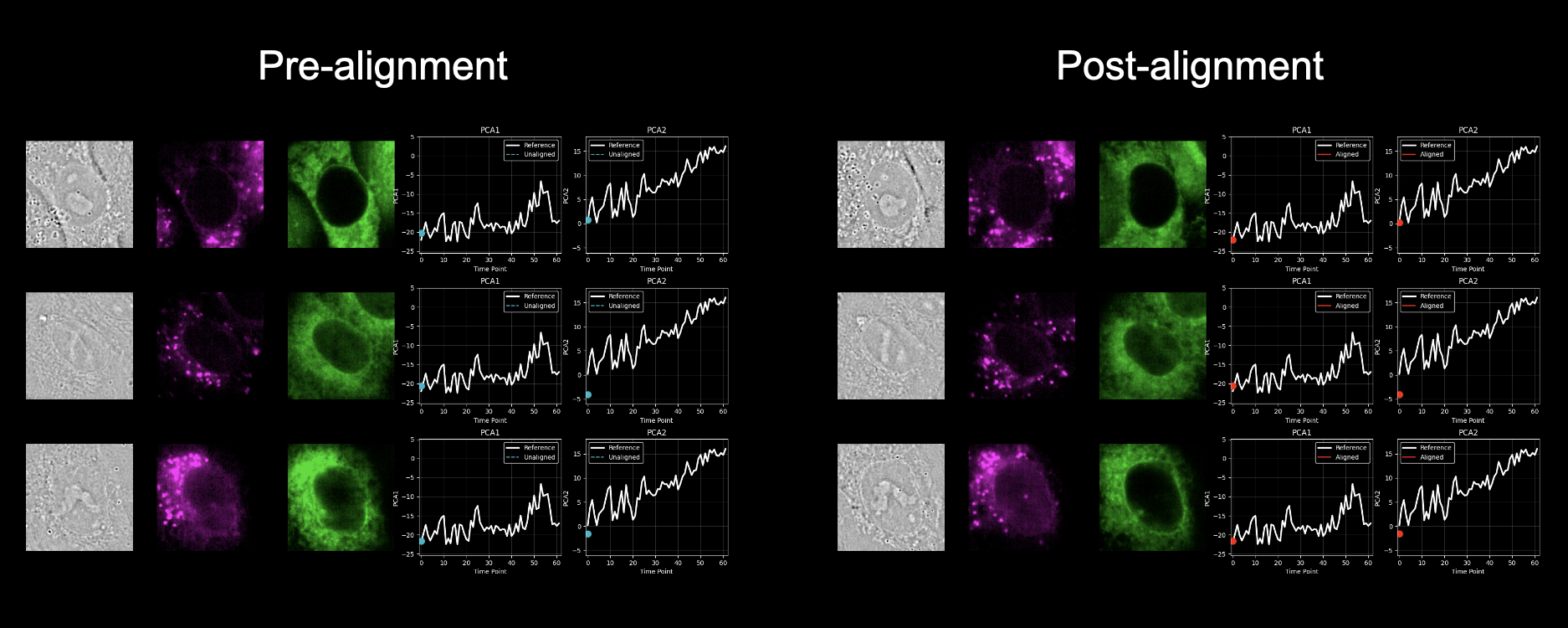}
\caption{\textbf{Alignment of the organelle remodeling response to infection}: The heterogenous response to infection can be synchronized via DTW using an annotated infection track. Principal component 1 and 2 displayed over time pre and post alignment.}
\label{vid:organelle_remodelling_DTW}
\end{Video}

\end{document}